%% file: emnlp2022.tex
\pdfoutput=1

\documentclass[11pt]{article}

\usepackage{EMNLP2022}

\usepackage{times}
\usepackage{latexsym}

\usepackage[T1]{fontenc}

\usepackage[utf8]{inputenc}

\usepackage{microtype}

\usepackage{inconsolata}

\usepackage{booktabs}
\usepackage{float}
\usepackage{comment}
\usepackage{todonotes}
\usepackage{xcolor,colortbl}
\usepackage{lipsum}  

\usepackage{amsmath,amssymb,amsthm,bm}
\usepackage{tikz-dependency}
\usepackage{tikz-qtree}
\usepackage{tikz}
\usepackage{stfloats}
\usetikzlibrary{shapes,arrows,chains}
\usepackage{multirow}
\usepackage{caption}
\usepackage{algorithm}
\usepackage{algpseudocode}
\algtext*{EndWhile} 
\algtext*{EndIf} 
\algtext*{EndFor} 
\algnewcommand{\Inputs}[1]{%
  \State \textbf{Inputs:}
  \Statex \hspace*{\algorithmicindent}\parbox[t]{.8\linewidth}{\raggedright #1}
}
\algnewcommand{\Output}[1]{%
  \State \textbf{Output:}
  \Statex \hspace*{\algorithmicindent}\parbox[t]{.8\linewidth}{\raggedright #1}
}
\algnewcommand{\Outputs}[1]{%
  \State \textbf{Outputs:}
  \Statex \hspace*{\algorithmicindent}\parbox[t]{.8\linewidth}{\raggedright #1}
}
\algnewcommand{\Compute}[1]{%
  \State \textbf{Compute}
  \Statex \hspace*{\algorithmicindent}\parbox[t]{.8\linewidth}{\raggedright #1}
}
\algnewcommand{\Initialize}[1]{%
  \State \textbf{Initialize:}
  \Statex \hspace*{\algorithmicindent}\parbox[t]{.8\linewidth}{\raggedright #1}
}
\algnewcommand{\OpEstimate}[1]{%
  \State \textbf{(Optional) Estimate:}
  \Statex \hspace*{\algorithmicindent}\parbox[t]{.8\linewidth}{\raggedright #1}
}
\algnewcommand{\Estimate}[1]{%
  \State \textbf{Estimate:}
  \Statex \hspace*{\algorithmicindent}\parbox[t]{.8\linewidth}{\raggedright #1}
}
\algnewcommand{\Infer}[1]{%
  \State \textbf{Infer:}
  \Statex \hspace*{\algorithmicindent}\parbox[t]{.8\linewidth}{\raggedright #1}
}

\usepackage{etoolbox}
\makeatletter
\patchcmd{\@verbatim}
  {\verbatim@font}
  {\verbatim@font\scriptsize}
  {}{}
\makeatother

\DeclareMathOperator*{\argmax}{arg\,max}

\DeclareMathOperator*{\pa}{pa}
\DeclareMathOperator*{\ch}{ch}
\DeclareMathOperator*{\graphroot}{root}
\DeclareMathOperator*{\sigmoid}{sigmoid}

\usepackage{xspace}
\usepackage{enumitem}
\usepackage{subfigure}

\newcommand{\smallsection}[1]{\noindent\textbf{#1.}}

\title{Neural-Symbolic Inference for Robust Autoregressive Graph Parsing via Compositional Uncertainty Quantification}

\newcommand\eqfootnote[1]{%
  \begingroup
  \renewcommand\thefootnote{}\footnote{#1}%
  \addtocounter{footnote}{-1}%
  \endgroup
}

\author{Zi Lin\\
  UC San Diego \\
  \texttt{lzi@ucsd.edu} \\\And
  Jeremiah Liu$^{\dagger\ddagger}$ \\
  Google Research \& Harvard University \\
  \texttt{jereliu@google.com} \\\And
  Jingbo Shang$^\dagger$ \\
  UC San Diego \\
  \texttt{jshang@ucsd.edu} \\
  }

\begin{document}
\maketitle
\eqfootnote{$^\dagger$ Co-senior authors. $^\ddagger$ Work done at Google.}
\begin{abstract}
Pre-trained seq2seq models excel at graph semantic parsing with rich annotated data, but generalize worse to out-of-distribution (OOD) and long-tail examples. 
In comparison, symbolic parsers under-perform on population-level metrics, but exhibit unique strength in OOD and tail generalization. 
In this work, we study compositionality-aware approach 
to neural-symbolic inference informed by model confidence, performing fine-grained neural-symbolic reasoning at subgraph level (i.e., nodes and edges) and precisely targeting subgraph components 
with high uncertainty in the neural parser. 
As a result, the method combines the distinct strength of the neural and symbolic approaches in capturing different aspects of the graph prediction, leading to well-rounded generalization performance both across domains and in the tail. We empirically investigate the approach in the English Resource Grammar (ERG) parsing problem on a diverse suite of standard in-domain and seven OOD corpora. 
Our approach leads to $35.26\%$ and $35.60\%$ error reduction in aggregated \textsc{Smatch} score over neural and symbolic approaches respectively, and $14\%$ absolute accuracy gain in key tail linguistic categories over the neural model, outperforming prior state-of-art methods that do not account for compositionality or uncertainty. 
\end{abstract}

\input{sections/01.introduction}
\input{sections/02.background}
\input{sections/03.uncertainty-calibration}
\input{sections/04.methods}

\input{sections/05.experiments}
\input{sections/06.related-work}
\input{sections/07.conclusion}

\section*{Acknowledgement}
Our work is sponsored in part by National Science Foundation Convergence Accelerator under award OIA-2040727 as well as generous gifts from Google, Adobe, and Teradata. Any opinions, findings, and conclusions or recommendations expressed herein are those of the authors and should not be interpreted as necessarily representing the views, either expressed or implied, of the U.S. Government. The U.S. Government is authorized to reproduce and distribute reprints for government purposes not withstanding any copyright annotation hereon.
We thank Du Phan, Panupong Pasupat, Jie Ren, Balaji Lakshminarayanan and Deepak Ramachandran for helpful discussion.

\section*{Limitation}
Here we discuss a potential limitations of the current study:

\paragraph{Problem domain} In this work, we have selected English Resource Grammar as the target formalism. This is a deliberate choice based on the availability of (1) realistic out-of-distribution evaluation corpus, and (2) well-established, high-quality symbolic parser. This is a common setting in industrial applications, where an practitioner is tempted to combine large pre-trained neural model with expert-developed symbolic rules to improve performance for a new domain. Unfortunately, we are not aware of another popular meaning representation for which both resources are available. To overcome this challenge, we may consider studying collaborative inference between a standard seq2seq model and some indirect symbolic supervision, e.g., syntactic parser or CCG parser \cite{steedman2001syntactic}, which is an interesting direction for future work.

\paragraph{Uncertainty estimation techniques} The vanilla seq2seq model is known to under-estimate the true probability of the high-likelihood output sequences, wasting a considerable amount of probability mass towards the space of improbable outputs \citep{ott2018analyzing, lebrun2022evaluating}. This systematic underestimation of neural likelihood may lead to a conservative neural-symbolic procedure that implicitly favors the information from the symbolic prior. It may also negatively impact calibration quality, leading the model to under-detect wrong predictions.
To this end, it is interesting to ask if a more advanced seq2seq uncertainty method (e.g., Monte Carlo dropout or Gaussian process \citep{gal2016dropout, liu2020simple}) can provide systematically better uncertainty quantification, and consequently improved downstream performance.

\paragraph{Graphical model specification} The GAP model presented in this work considers a classical graphical model likelihood $p(G|x)=\prod_{v\in G} p(v|\pa(v), x)$ , which leads to a clean factorization between graph elements $v$ and fast probability computation. However, it also assumes a local Markov property that $v$ is conditional independent to its ancestors given the parent $\pa(v)$. In theory, the probability learned by a seq2seq model is capable of modeling higher order conditionals between arbitrary elements on the graph. Therefore it is interesting to ask if a more sophisticated graphical model with higher-order dependency structure can lead to better performance in practice while maintaining reasonable computational complexity.

\paragraph{Understanding different types of uncertainty} There exists many different types of uncertainties occur in a machine learning system \citep{hullermeier2021aleatoric}. This includes data uncertainty (e.g., erroneously annotated training labels, ill-formedness of the input sentence, or inherent ambiguity in the example-to-label mapping), and also model uncertainty which occurs the test example not containing familiar patterns the model learned from the training data. In this work, we quantifies uncertainty using mean log likelihood, which broadly captures both types of uncertainty and does not make a distinction between these different subtypes. As different source of uncertainty may lead to different strategy in neural-symbolic parsing, the future work should look into more fine-grained uncertainty signal that can decompose these different sources of error and uncertainty, and propose adaptive strategy to handle different scenarios.


\section*{Ethical Consideration}
This paper focused on neural-symbolic semantic parsing for the English Resource Grammar (ERG). Our architecture are built based on open-source models and datasets (all available online). We do not anticipate any major ethical concerns.

\bibliography{anthology,custom}
\bibliographystyle{acl_natbib}

\appendix
\input{sections/appendices/meaning-representation}
\input{sections/appendices/graph-semantic-parsing}
\input{sections/appendices/additional-discussion}
\input{sections/appendices/extension-implementation}
\input{sections/appendices/graph-alignment-algorithm}

\input{sections/appendices/ood-datasets}
\input{sections/appendices/implementation}

\input{sections/appendices/ind-test}
\input{sections/appendices/linguistic-phenomena}
\input{sections/appendices/calibration}

\end{document}

%% file: sections/01.introduction.tex
\section{Introduction}
A structured account of compositional meaning has become a longstanding goal for Natural Language Processing. To this end, a number of efforts have focused on encoding semantic relationships and attributes into graph-based meaning representations (MRs, see Appendix \ref{app:meaning-representation} for details). 
In particular, graph semantic parsing has been an important task in almost every Semantic Evaluation (SemEval) exercise since 2014.
In recent years, we have witnessed the burgeoning of applying neural networks to semantic parsing. Pre-trained language model-based approaches have led to significant improvements across different MRs~\citep{oepen-etal-2019-mrp,oepen-etal-2020-mrp}.
However, these models often generalize poorly to out-of-distribution (OOD) and tail examples~\citep{cheng-etal-2019-learning,shaw-etal-2021-compositional,kim2021sequence,lin-etal-2022-towards}, while grammar or rule-based parser work relatively robustly across different linguistic phenomena and language domains~\cite{cao-etal-2021-comparing,lin-etal-2022-towards}. See Section \ref{sec:related-work} for a review of related work.


In this paper, we propose a novel compositional neural-symbolic inference for graph semantic parsing, which takes advantage of both uncertainty quantification from a seq2seq parser and prior knowledge from a symbolic parser at the subgraph level (i.e., nodes and edges).
We take graph semantic parsing for English Resource Grammar (ERG) as our case study. 
ERG is a compositional semantic representation explicitly coupled with the syntactic structure. 
Compared to other graph-based meaning representations like Abstract Meaning Representation (AMR), ERG has high coverage of English text and strong transferability across domains, rendering itself as an attractive target formalism for automated semantic parsing. Furthermore, many years of ERG research has led to well-established symbolic parser and a rich set of carefully constructed corpus across different application domains and fine-grained linguistic phenomena, 
making it an ideal candidate for studying cross-domain generalization of neural-symbolic methods
\citep{oepen-etal-2002-lingo, crysmann-packard-2012-towards}.

We start with a novel investigation of the uncertainty calibration behaviour of a T5-based state-of-the-art neural ERG parser~\citep{lin-etal-2022-towards} on the subgraph level (Section~\ref{sec:calibration}), where we make some key observations: 
(1) the performance of the neural parser degrades when it becomes uncertain at the subgraph level, while 
(2) the symbolic parser works still robustly when the neural parser is uncertain at the subgraph level. This motivates us to develop a \textit{compositional} neural-symbolic inference process where the neural and symbolic parser collaborates at a more fine-grained level and guided by model uncertainty, which is an aspect missing in the previous neural-symbolic and ensemble parsing literature (see Appendix \ref{sec:related-work}).


\input{figures/example}

We then propose a decision-theoretic criteria to allow for neural-symbolic inference at subgraph level (i.e., nodes and edges) and incorporates the neural parser's fine-grained uncertainty for each graph component prediction (Section \ref{sec:compositional-inference}).
The key to this approach is a \textit{meta graph} $\mathcal{G}_M$ that enumerates possible candidates for each node/edge prediction, and is constructed by merging multiple beam predictions from the neural seq2seq model.

The core challenge here is how to properly quantify \textit{compositional uncertainty} using a seq2seq model, i.e., assigning model probability for a node or edge prediction.
For example, our interest is to express the conditional probability of a graph node $v$ with respect to its parent $p(v|pa(v), x)$, rather than the likelihood of $v$ conditioning on the previous tokens in the linearized string. 
As a result, 
it cannot be achieved by relying on the naive token-level autoregressive probabilities from the beam search.
To address this issue, we introduce a simple probabilistic formalism termed \textit{Graph Autoregressive Process} (GAP) (Section \ref{sec:compositional-uncertainty}). GAP adopts a dual representation of an autoregressive process and a probabilistic graphical model, and can serve as a powerful medium for expressing compositional uncertainty for seq2seq graph parsing.

We demonstrate the effectiveness of our approach in experiments across a diverse suite of eight in-domain and OOD evaluation datasets encompassing 
domains including Wikipedia entries, news articles, email communications, etc (Section \ref{sec:exp}). We achieve the best results on the overall performance across the eight domains, attaining $35.26 \%$ and $35.60 \%$ error reduction in the aggregated \textsc{Smatch} score over the neural and symbolic parser, respectively. 
Our approach also exhibits significantly stronger robustness in generalization to OOD datasets and long-tail linguistic phenomena than previous work, while maintaining the state-of-the-art performance on in-domain test. Further study also shows that the compositionality aspects of neural-symbolic inference helps the model to assemble novel graph solution that the original inference process (e.g., beam search or symbolic parse) fails to provide (Section \ref{sec:case-study}).

In summary, our contributions are four-fold:
\begin{itemize}[nosep,leftmargin=*]
    \item We present a novel investigation of neural graph parser's uncertainty calibration performance at \textit{subgraph level} (Section \ref{sec:calibration}). Our study confirms the seq2seq uncertainty is effective for detecting model error even out-of-distribution, establishing the first empirical basis for the utility of \textit{compositional} uncertainty in seq2seq graph parsing.
    \item We propose a practical and principled framework for neural-symbolic graph parsing that utilizes model uncertainty and exploits compositionality (Section \ref{sec:compositional-inference}). The method is fully compatible with modern large pre-trained seq2seq network using beam decoding, and is 
    general-purpose and applicable to any graph semantic parsing task. 
    \item We propose a simple probabilistic formalism (GAP) to express a seq2seq model's compositional uncertainty (Section \ref{sec:compositional-uncertainty}). 
    GAP allows us to go beyond the conventional autoregressive sequence probability and express long-range parent-child conditional probability on the graph, serving as a useful medium of compositional uncertainty quantification.
    \item We conduct a comprehensive study to evaluate the state-of-the-art graph parsing approaches across a diverse suite of in-domain and out-of-distribution datasets (Section \ref{sec:exp}). Our study reveals surprising weakness of previous neural-symbolic methods in OOD generalization, and confirms the proposed method significantly improves models OOD and tail performance. 
\end{itemize}
\smallsection{Reproducibility} Our code is available on Github:~\url{https://github.com/google/uncertainty-baselines/tree/main/baselines/t5/data/deepbank}.



%% file: figures/example.tex
\begin{figure*}[!ht]
 \centering
 \subfigure[EDS Representation]{
    \includegraphics[width=1.1\columnwidth]{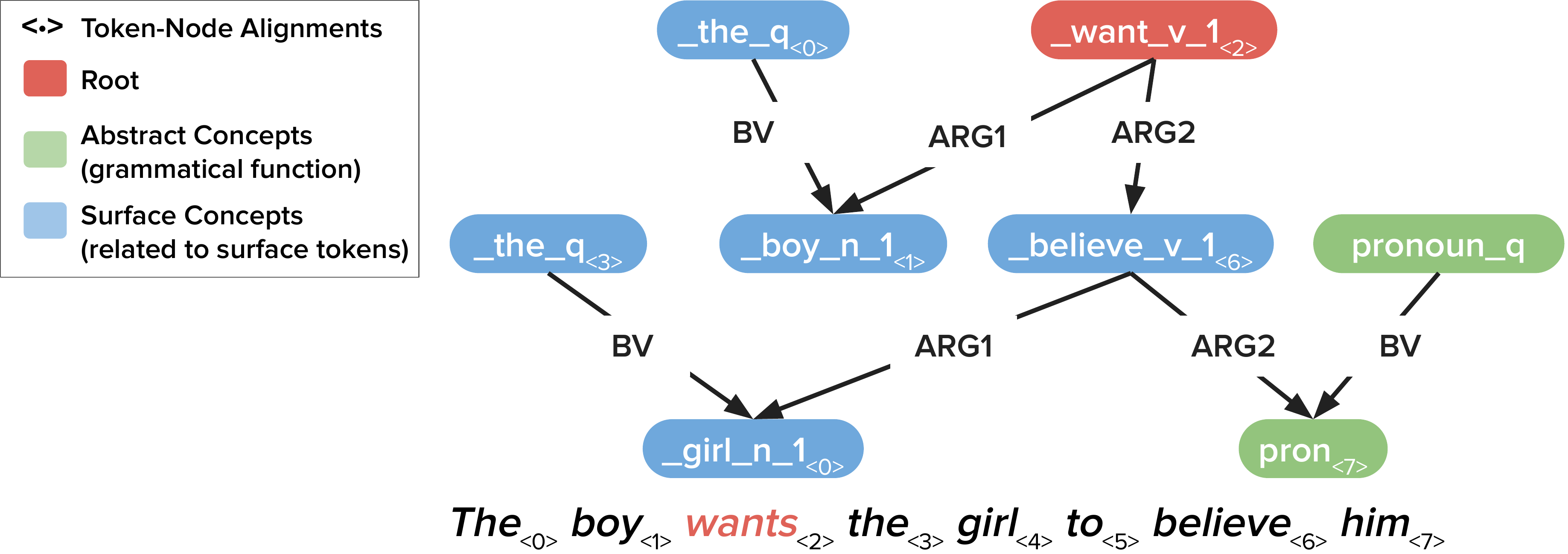}
    \label{fig:erg-exp}
 }
 \hspace{1.5em}
 \subfigure[Variable-free PENMAN notation]{
    \includegraphics[width=0.35\linewidth]{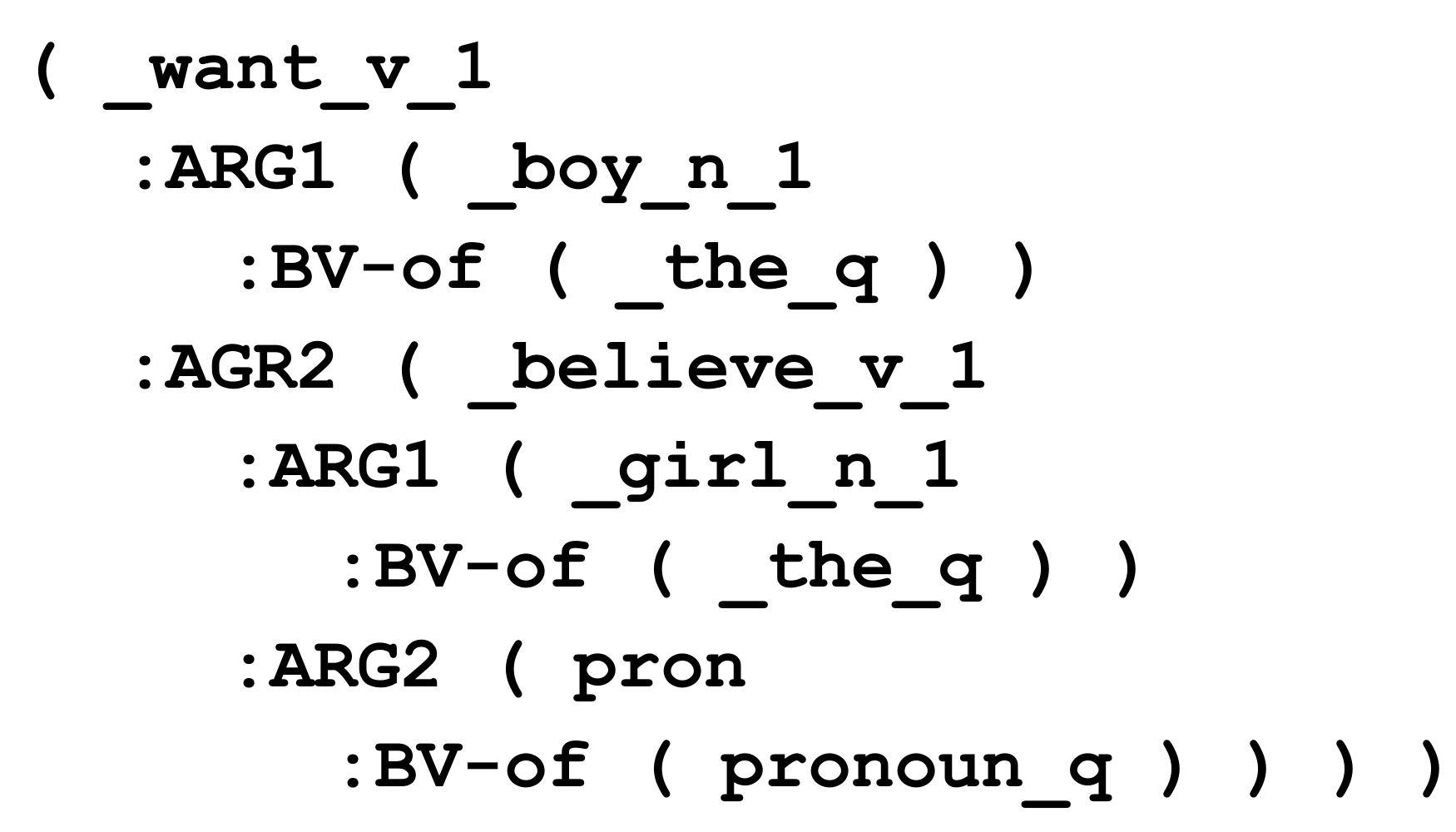}
    \label{fig:penman-exp}
}
\vspace{-1em}
\caption{
The EDS representation for ERG and the corresponding linearization of the example sentence ``\textit{The boy wants the girl to believe him}''.
}
\vspace{-1.5em}
\end{figure*}

%% file: sections/02.background.tex
\input{figures/calibration}

\section{Background}
\label{sec:background}
\subsection{English Resource Grammar (ERG)}
In this work, we take the representations from English Resource Grammar \citep[ERG;][]{flickinger-etal-2014-towards} as our target meaning representations.
ERG is a broad-coverage computational grammar of English that derives underspecified logical-form representations of meaning \citep{oepen-flickinger-2019-erg}. It is rooted in the general linguistic theory of Head-driven Phrase Structure Grammar \citep[HPSG;][]{pollard1994head}. 


ERG can be presented into different types of annotation formalism \citep{copestake2005minimal}. This work focuses on the Elementary Dependency Structure \citep[EDS;][]{oepen-lonning-2006-discriminant} which is a compact representation that can be expressed as a directed acyclic graph (DAG) and is widely adopted in the neural parsing approaches \citep{buys-blunsom-2017-robust,chen-etal-2018-accurate}. An example is shown in Figure \ref{fig:erg-exp}.



\subsection{Parsing Approaches}
\label{sec:parsing-to-erg}
In this section, we review the state-of-the-art symbolic and neural parsers utilized in our work, i.e., the ACE parser \citep{crysmann-packard-2012-towards} and the T5 parser \citep{lin-etal-2022-towards}. Appendix \ref{app:gsp-review} reviews other ERG parsing techniques.


\smallsection{The symbolic parser: ACE} 
The ACE parser \citep{crysmann-packard-2012-towards} is one of the state-of-the-art symbolic parsers.
It first decomposes sentences into ERG-consistent candidate derivation trees, and the parser will rank candidates based on the structural features in the nodes of the derivation trees via maximum entropy models \citep{oepen-lonning-2006-discriminant, toutanova2005stochastic}. This approach fails to parse sentences for which no valid derivation is found.


\smallsection{The neural parser: T5} \citet{lin-etal-2022-towards} proposed a T5-based ERG parser which achieves the best known results on the in-domain DeepBank benchmark. It is the first work that successfully transfers the ERG parsing problem into a pure end-to-end translation problem via compositionality-aware tokenization and a variable-free top-down graph linearization 
based on the PENMAN notation \citep{kasper1989flexible}.
Figure \ref{fig:penman-exp} shows an example of the linearized graph string from the original EDS graph.

%% file: figures/calibration.tex
\begin{figure*}[ht]
 \centering
 \subfigure{
    \includegraphics[width=0.46\linewidth]{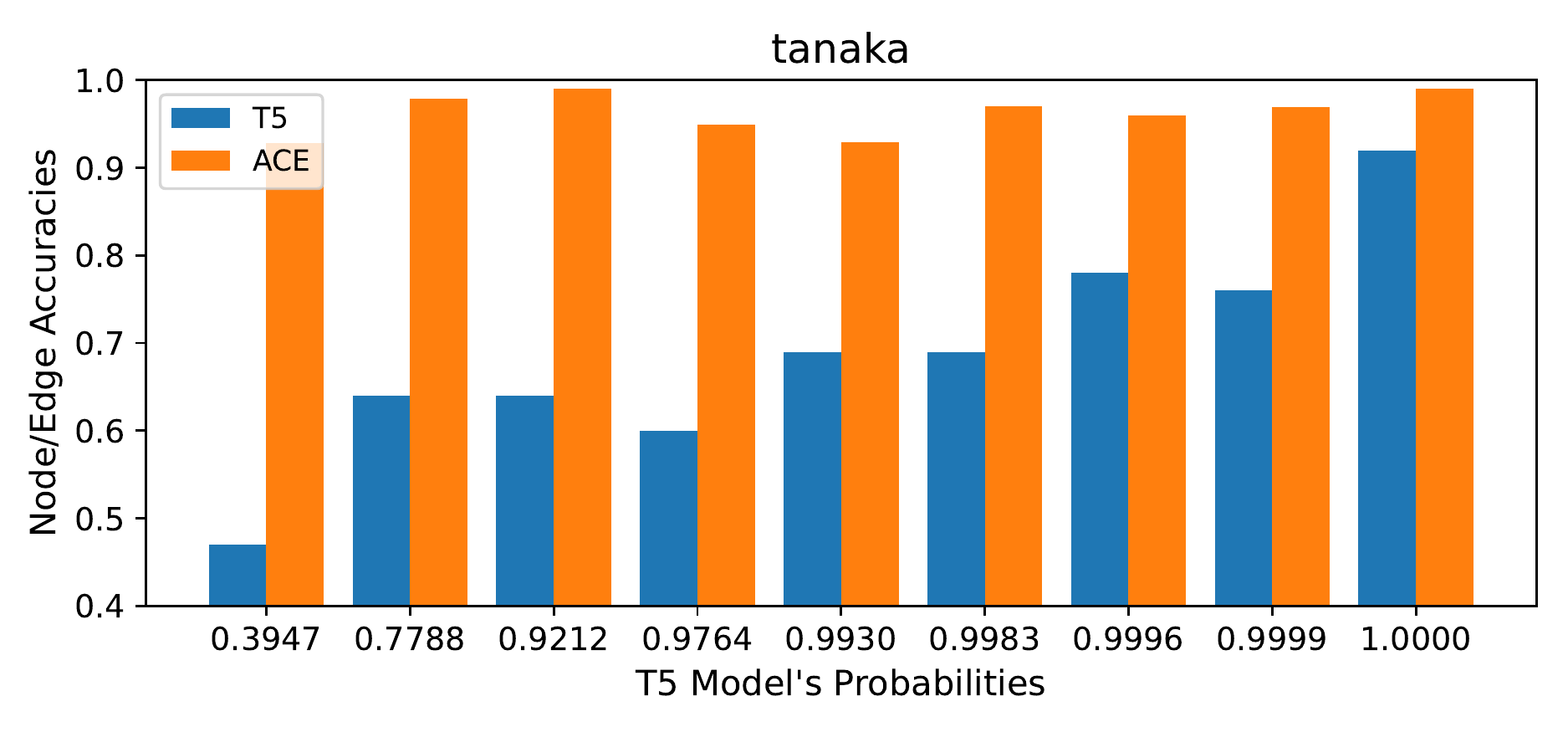}
 }
 \subfigure{
    \includegraphics[width=0.46\linewidth]{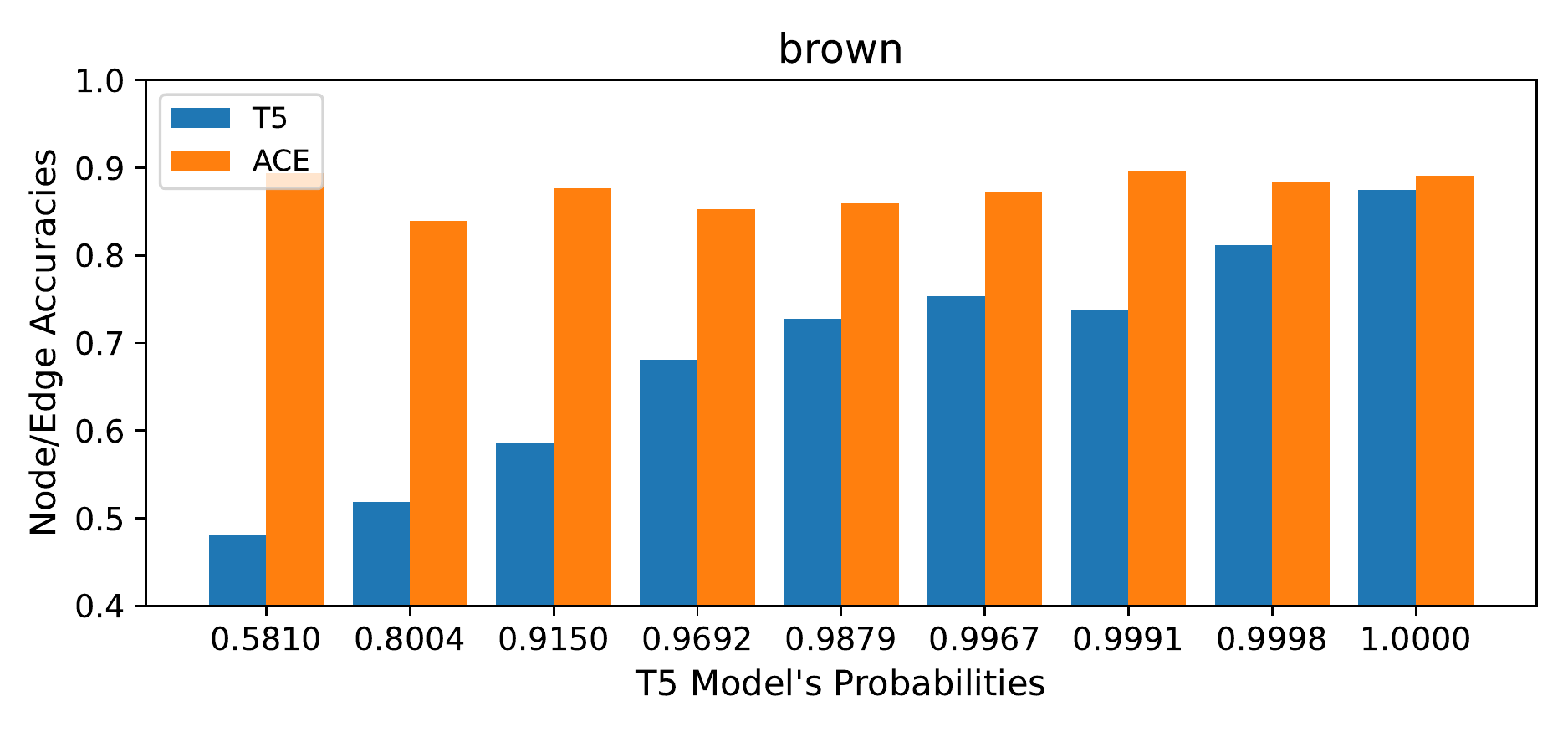}
}
\vspace{-1.5em}
\caption{Bar charts for the predictive accuracies of the T5 parser (blue) and ACE parser (orange) for all the node / edge prediction across different uncertainty buckets based on T5 model's probabilities. The performance is evaluated on the Tanaka and Brown datasets. Each bin represents a quantile bucket of the model probability (i.e., they contain the same number of examples). Since at most of the subgraphs, the model is pretty certain ($\log P > -1e-5$), we exclude these pretty certain predictions in the figures.
\vspace{-1em}
}
\label{fig:calibration}
\end{figure*}

%% file: sections/03.uncertainty-calibration.tex
\section{Motivation: Subgraph-level Uncertainty in Seq2seq Graph Parsing}
\label{sec:calibration}

We hypothesize that when the neural seq2seq model is uncertain at the subgraph level, it is more likely to make mistakes. 
Assuming the symbolic parser performs more robustly in these situations, we can then design a procedure to ask the symbolic parser for help when the model is uncertain. To validate this hypothesis, we conduct experiments to empirically explore the following two questions: (1) how does the model perform when it is uncertain at the subgraph level? and (2) how does the symbolic parser perform when the model is uncertain?

First, we compute model probabilities for each graph element (i.e., node and edge) prediction (see Section \ref{sec:compositional-uncertainty} for how to compute these quanitities), and identify the corresponding ACE parser prediction using the graph matching algorithm from \textsc{Smatch}~\citep{cai-knight-2013-smatch}. 
We then evaluate the accuracies of those graph element predictions with respect to the gold labels, and compare it to that of the ACE parser.

In Figure~\ref{fig:calibration}, we plot the bar charts compare the neural and symbolic performance in different bucket of seq2seq model uncertainties on the two largest datasets (e.g., Tanaka and Brown, see Appendix~\ref{app:ood-datasets}). Results on other datasets can be found in the Appendix~\ref{app:calibration}.
As shown in the figure, low model probability generally corresponds to low T5 performance, while the corresponding ACE parser’s accuracies spread relatively stably (e.g., it attains $>90\%$ accuracy in the lowest-confidence buckets, while T5 accuracy is $<50\%$).
This implies that when the model is uncertain, the accuracy of the neural model tend to be low, while the ACE parser still performs well. This has motivated us to develop a \textit{compositional} neural-symbolic inference procedure guided the model's subgraph level uncertainty, such that the T5 and ACE parser can collaborate at a more fine-grained level via \textit{compostional uncertainty quantification} (Section~\ref{sec:methods}).

%% file: sections/04.methods.tex
\section{Methods}
\label{sec:methods}



\smallsection{Notation \& Problem Statement} For graph semantic parsing, the input is a natural language utterance $x$, and the output is a directed acyclic graph (DAG) $G = \langle \mathbf{N}, \mathbf{E} \rangle$, where $\mathbf{N}$ is the set of nodes and $\mathbf{E}\in \mathbf{N}\times \mathbf{N}$ is the set of edges (e.g., Figure \ref{fig:erg-exp}). In the case of seq2seq parsing, $G$ is represented as a linearized graph string $g=s_1s_2\cdots s_L$ which consists of symbols $\{s_l\}_{l=1}^L$ (e.g., Figure \ref{fig:penman-exp}). As the graph prediction is probabilistic, each of the graph element $v \in \textbf{N} \cup \textbf{E}$ is a random variable whose values are the symbols $s_i$ observed from the beam outputs, leading to marginal probabilities $p(v=s_i|x)$ and conditional probabilities $p(v=s_i| v'=s_j, x)$.

To this end, our goal is to produce a principled inference procedure for graph prediction 
accounting for model uncertainty on predicting graph elements $v \in G$.
In the sequel, Section \ref{sec:compositional-inference} presents a decision-theoretic criterion that leverages the graphical model likelihood $p(G|x)$ to conduct compositional neural-symbolic inference for graph prediction. To properly express the graphic model likelihood $p(G|x)=\prod_{v\in G} p(v|pa(v),x)$ using a learned seq2seq model, Section \ref{sec:compositional-uncertainty} introduces a simple probabilistic formalism termed \textit{Graph Autoregressive Process} (GAP) to translate the autoregressive sequence probability from the seq2seq model to graphical model probability. Appendix \ref{app:extension} discusses some additional extensions.

\input{sections/subsections/04.02.CNSI}
\input{sections/subsections/04.01.GAP}

%% file: sections/subsections/04.02.CNSI.tex
\subsection{Compositional Neural-Symbolic Inference}
\label{sec:compositional-inference}
Previously, an uncertainty-aware decision criteria was proposed for neural-symbolic inference based on the Hurwicz pessimism-optimism criteria $R(G|x)$ \citep{lin-etal-2022-towards}. Specifically, the criteria is written as:\\ 
$$R(G|x) =\alpha(x) * R_p(G|x) + (1-\alpha(x)) * R_0(G),$$
where $R(G|x)=-\log p(G|x)$ is the neural model likelihood, $R_0(G) = \log p_0(G)$ is the symbolic prior likelihood,
and 
$\alpha(x)$
is a the uncertainty-driven trade-off coefficient to balance between the optimistic MLE criteria $R_p(G|x)$ and the pessimistic, prior-centered criteria $R_0(G|x)$ centered around symbolic prediction $G_0$. 

A key drawback of this approach is the lack of accounting for the compositionality. 
This motivates us to consider synthesizing the 
multiple graph predictions $\{G_k\}_{k=1}^K$ from the neural parser to form a \textit{meta graph} $\mathcal{G}$ \footnote{Given a group of candidate graphs $\{G_k\}_{k=1}^K$, well-established algorithm exists to synthesize different graph predictions into a \textit{meta} graph $\mathcal{G}$ \citep{cai-knight-2013-smatch,hoang2021ensembling} (see Appendix \ref{app:graph-align} for a more detailed review). }, where we can leverage the disentangled uncertainty of $p(G|x)$ to perform fine-grained neural-symbolic inference for each graph component $v \in G$ (i.e., nodes or edges).
Specifically, 
we leverage the factorized graphical model likelihood $p(G|x) = \prod_{v\in G} p(v|\pa(v), x)$ to decompose the overall decision criteria $R(G|x)$ into that of individual components $R(v|x)$:
\begin{align}
    R(v|x) &= \alpha(v|x) * \log p(v|\pa(v), x)
    \nonumber \\
    &+ (1-\alpha(v|x))* \log p_0(v),
    \label{eq:criteria}
\end{align}
and the overall criteria is written as $R(G|x)=\sum_{v\in G} R(v|x)$. Here $\pa(v)$ refers to the parents of $v$ in $G$, and $\alpha(v|x)=\sigmoid(-\frac{1}{T}H(v|x)+b)$ is the component-specific trade-off parameter driven by model uncertainty $H(v|x)=-\log p(v|\pa(v), x)$, and $(T, b)$ are scalar calibration hyperparameters that can be tuned on the dev set. 

Following previous work \cite{lin-etal-2022-towards}, the symbolic prior $p_0$ for each graph component $v$ is defined as a Boltzmann distribution based on the graph output $G_0$ from the symbolic parser,  i.e., $p_0(v=s) \propto \exp(I(s \in G_0))$, so that it is proportional to the empirical probability of whether a symbol $s$ appears in $G_0$.
Notice that we have ignored the normalizing constants since they do not impact optimization.

Algorithm \ref{alg:CNSI} summarizes the full algorithm. As shown, during inference, the method proceeds by starting from the root node $v_0$ and selects the optimal prediction $\hat{v}_0=\argmax_{c_0 \in \text{Candidate}(v_0)}R(c_0|x)$, where $c_0$ are different candidates for $v_0$ given by the \textit{meta graph}  $\mathcal{G}$. The algorithm then recursively performs the same neural-symbolic inference procedure for the children of $v_0$ (i.e., $\ch(v)$). The algorithm terminates when the optimal candidates for all graph variables $v \in G$ are determined.

As a result, the algorithm is able to adaptively combine subgraph predictions across multiple beam candidates thanks to the meta graph $\mathcal{G}$, and appropriately weight between the local neural and symbolic information thanks to the uncertainty-aware decision criteria $R(v|x)$. Empirically, this also gives the algorithm the ability to synthesize novel graph predictions that are distinct from its base models (Section \ref{sec:case-study}).

\input{algorithm/algorithm2}

%% file: algorithm/algorithm2.tex
\begin{algorithm}[ht]
\captionsetup{font=small}
\caption{Compositional Neural-Symbolic Inference}\label{alg:CNSI}
\small
\begin{algorithmic}
\Inputs{Meta graph $\mathcal{G}$\\Graphical model likelihood $\log p(G|x)$\\Symbolic prior $p_0$}
\Output{Neural-symbolic graph prediction $G$}
\Initialize{$v = \graphroot(G_M)$; $G = \mathcal{G}_M$.}
\If{$G$ does not contain undecided candidates}
\Return{$G$}
\Else{}
\For{$c_v \in \text{Candidate}(v)$}
    \State{Compute decision criteria $R(c_v|x)$ (Equation \ref{eq:criteria})}
\EndFor
\State{Select optimal candidate $\hat{v}=\argmax_c R(c|x)$}
\State{Remove non-optimal candidates of $v$ from $G$}
\State{Recursively perform Algorithm \ref{alg:CNSI} for all $v' \in \ch(v)$}
\EndIf
\end{algorithmic}
\end{algorithm}
\vspace{-1em}


%% file: sections/subsections/04.01.GAP.tex
\subsection{Compositional Uncertainty Quantification with Graph Autogressive Process (GAP)}
\label{sec:compositional-uncertainty}
To properly model the uncertainty $p(G|x)$ from a seq2seq model, we need an intermediate probabilistic representation to translate the raw token-level probability to the distribution over graph elements.

To this end, we introduce a simple probabilistic formalism termed \textit{Graph Autoregressive Process} (GAP), which is a probability distribution assigning seq2seq learned probability to the graph elements $v \in G$.
Specifically, as the seq2seq-predicted graph adopts both a sequence-based representation $g=s_1, ..., s_L$ and a graph representation $G = \langle \mathbf{N}, \mathbf{E} \rangle$, the GAP model adopts both an autoregressive representation $p(g|x) = \prod_{i} p(s_i|s_{<i}, x)$ (Section \ref{sec:g-GAP}), and also a probabilistic graphical model representation $p(G|x) = \prod_{v\in G} p(v|\pa(v), x)$ (Section \ref{sec:G-GAP}). Both representations share the same set of underlying probability measures (i.e., the graphical-model likelihood $p(G|x)$ can be derived from the autoregressive probabilities $p(s_i|s_{<i}, x)$) (Figure \ref{fig:G-GAP}), rendering itself a useful medium for principled compositional neural-symbolic inference using seq2seq probabilities.


\subsubsection{Autoregressive Representation for Linearized Sequence $g$}
\label{sec:g-GAP}
Given an input sequence $x$ and output sequence $y = y_1y_2\cdots y_N$, the token-level autoregressive distribution from a seq2seq model is $p(y|x) = \prod_{i=1}^N p(y_i|y_{<i}, x)$.
In the context of graph parsing, the output sequence describes a linearized graph $g=s_1s_2\cdots s_L$, where each symbol $s_i=\{y_{i_1}y_{i_2}\cdots y_{i_{N_i}}\}$ represents either a node $n \in \mathbf{N}$ or an edge $e\in \mathbf{E}$ of the graph and corresponds to a collection of beam-decoded tokens $\{y_{i_1}y_{i_2}\cdots y_{i_{N_i}}\}$, e.g., the node \texttt{\_the\_q} in Figure \ref{fig:erg-exp} is represented by tokens \{\texttt{\_}, \texttt{the}, \texttt{\_q}\}.
This process is illustrated in follows:
\vspace{-0.5em}
\begin{figure}[H]
    \centering
    \includegraphics[width=0.65\columnwidth]{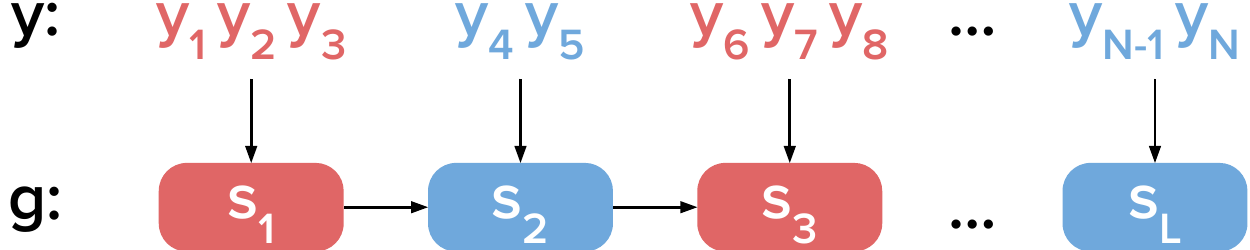}
\end{figure}
\vspace{-0.5em}
To this end, the \textit{Graph Autoregressive Process} (GAP) assigns probability to each linearized graph $g=s_1s_2\cdots s_L$ autoregreesviely as $p(g|x) = \prod_{i=1}^L p(s_i|s_{<i}, x)$, and the conditional probability $p(s_i|s_{<i}, x)$ is computed by aggregating the token probability:
\resizebox{\linewidth}{!}{
\begin{minipage}{\linewidth}
{\footnotesize
\begin{align*}
    p(s_i|s_{<i}, x) 
    &= p(\{y_{i_1}\cdots y_{i_{N_i}}\}|s_{<i}, x)
    = \prod_{j=1}^{N_i} p (y_{i_j}|y_{i_{<j}}, s_{<i}, x)
\end{align*}
}%
\end{minipage}
}
\smallsection{Marginal and Conditional Probability} Importantly, GAP allows us to compute the marginal and (non-local) conditional probabilities for graph elements $s_i$. Given the input $x$, the marginal probability of $s_i$ is computed as
\resizebox{\linewidth}{!}{
\begin{minipage}{\linewidth}
\begin{align*}
    p(s_i|x) = \int_{s_{<i}} p(s_i|s_{<i}, x)p(s_{<i}|x)\mathrm{d}s_{<i}
\end{align*}
\end{minipage}
}
by integrating over the space of all possible subsequences $s_{<i}$ prior to the symbol $s_i$. Then, the (non-local) conditional probability between two graph elements $(s_i, s_j)$ with $i < j$ is computed as
\resizebox{\linewidth}{!}{
\begin{minipage}{\linewidth}
{\footnotesize
\begin{align*}
    &p(s_j|s_i, x) = \\
    &\int_{s_{i\rightarrow j}, s_{<i}} 
    p(s_i, s_{i\rightarrow j}|s_i, s_{<i}, x) 
    p(s_i|s_{<i}, x) p(s_{<i}|x) 
    \mathrm{d}s_{i\rightarrow j} \mathrm{d}s_{<i}
\end{align*}
}%
\end{minipage}
}
by integrating over the space of subsequences $s_{i\rightarrow j}$ between $(s_i, s_j)$ and the subsequence $s_{<i}$ before $s_i$. Higher order conditional (e.g., $p(s_j|(s_i, s_l), x)$) can be computed analogously. Notice this gives us the ability to reason about long-range dependencies between non-adjacent symbols on the sequence. Furthermore, the conditional probability on the \textit{reverse} direction can also be computed using the Bayes' rule: $p(s_i|s_j, x) = \frac{p(s_j|s_i, x)p(s_i|x)}{p(s_j|x)}$.

\smallsection{Efficient Estimation Using Beam Outputs} In practice, we can estimate $p(s_i|x)$ and $p(s_j|s_i, x)$ efficiently via  importance sampling using the output from the beam decoding $\{g_k\}_{k=1}^K$, where $K$ is the beam size \citep{malinin2020uncertainty}. The marginal probability can be computed as 
\resizebox{\linewidth}{!}{
\begin{minipage}{\linewidth}
\begin{align}
\label{eq:marginal-prob}
    \hat{p}(s_i|x) = \sum_{k=1}^K \pi_k p(s_i|s_{k, <i}, x)
\end{align}
\end{minipage}
}
where $\pi_k = \frac{\exp(\frac{1}{t} \log p (g_k|x))}{\sum_{k=1}^K \exp(\frac{1}{t} \log p (g_k|x))}$ is the importance weight proportional to the beam candidate $g_k$'s log likelihoods, and $t>0$ is the temperature parameter fixed to a small constant (e.g., $t=0.1$, see Appendix \ref{app:prob-estimation} further  discussion) \citep{malinin2020uncertainty}. 
If the symbol $s_i$ does not appear in the $k^{th}$ beam, we set $p(s_i|s_{k, <i}, x) = 0$.

Then, for two symbols $(s_i, s_j)$ with $i<j$, we can estimate the joint probability as
\vspace{-0.5em}

\resizebox{\linewidth}{!}{
\begin{minipage}{\linewidth}
{\footnotesize
\begin{align}
\label{eq:conditional-prob}
    \hat{p}(s_j | s_i, x) = \sum_{k=1}^K \pi_k^i p(s_j|s_i, s_{k, i\rightarrow j}, s_{k, <i}, x)
\end{align}
}%
\end{minipage}
}
where $\pi_k^i = \frac{\exp(\frac{1}{t} \log p (g_k|x)) * I(s_i \in g_k)}{\sum_{k=1}^K \exp(\frac{1}{t} \log p (g_k|x))  * I(s_i \in g_k)}$ is the importance weight among beam candidates that contains $s_i$. Notice this is different from Equation \ref{eq:marginal-prob} where $\pi_k$ is computed over all beam candidates regardless of whether it contains $s_i$.

\subsubsection{Probabilistic Graphical Model Representation for $G$}
\label{sec:G-GAP}

So far, we have focused on probability computation based on the graph's linearized representation $p(g|x)=\prod_i p(s_i|s_{<i}, x)$.
To conduct the compositional neural-symbolic inference (Section \ref{sec:compositional-inference}), we also need to consider GAP's graphical model representation $p(G|x) = \prod_{v\in G} p(v|\pa(v), x)$.


\input{figures/G-GAP}

GAP's graphical model representation $G$ depends on the \textit{meta graph} $\mathcal{G}$ constructed from $K$ candidate graphs $\{G_k\}_{k=1}^K$ (Section \ref{sec:compositional-inference}).
Figure \ref{fig:G-GAP} shows an example, where $n_i$ and $e_j$ are the candidates for the node and edge predictions collected from beam sequences. Compared to the sequence-based representation $g$, $\mathcal{G}$ provides two advantages: it (1) explicitly enumerates different candidates for each node and edge prediction (e.g., $n_2$ v.s. $n_3$ for predicting the third element), and (2) provides an explicit account of the parent-child relationships between variables on the graph (e.g., $e_2$ is a child node of $n_1$ in the predicted graph, which is not reflected in the autoregressive representation). From the probabilistic learning perspective, $\mathcal{G}$ describes the space of possible graphs (i.e., the \textit{support}) for a graph distribution $p(G|x): G \rightarrow [0, 1]$. 


To this end, GAP assigns proper graph-level probability $p(G|x)$ to graphs $G$ sampled from the meta graph $\mathcal{G}$ via the graphical model likelihood:
\resizebox{\linewidth}{!}{
\begin{minipage}{\linewidth}
\begin{align*}
    p(G|x) &= \prod_{v\in G} p(v|\pa(v), x) \\
          &= \prod_{n\in \mathbf{N}} p(n|\pa(n), x) * \prod_{e\in \mathbf{E}} p(e|\pa(e), x)
\end{align*}
\end{minipage}
}
where $p(v|\pa(v), x)$ is the conditional probability for $v$ with respect to their parents $\pa(v)$ in $G$. Given the candidates graphs $\{G_k\}_{k=1}^K$, we can express the likelihood for $p(v|\pa(v), x)$ by writing down a multinomial likelihood enumerating over different values of $\pa(v)$ \citep{murphy2012machine}.
This in fact leads to a simple expression for the model likelihood as a simple averaging of the beam-sequence log likelihoods:
\resizebox{\linewidth}{!}{
\begin{minipage}{\linewidth}
{\footnotesize
\begin{align}
    & \log p(n|\pa(n), x)
    \propto \frac{1}{K} \sum_{k=1}^K \log p(n|\pa(n)=c_k)
    \label{eq:graphical-model-likelihood}
\end{align}
}%
\end{minipage}
}
where $c_k$ is the value of $\pa(n)$ in $k^\text{th}$ beam sequence, and the conditional probabilities are computed using Equation (\ref{eq:conditional-prob}). See Appendix \ref{app:likelihood-computation} for a detailed derivation.

\input{algorithm/algorithm1}

In summary, for each graph element variable $v \in G$, GAP allows us to compute the graphical-model conditional likelihood $p(v|pa(v), x)$ via its graphical model representation, and also to compute the marginal probability $p(v|x)$ via its autoregressive presentation. The conditional likelihood is crucial for neural-symbolic inference (Section \ref{sec:compositional-inference}), and the marginal probability is useful for sparsity regularization in global graph structure inference (Appendix \ref{app:extension}). Algorithm \ref{alg:GAP} summarizes the full GAP computation.

%% file: figures/G-GAP.tex
\begin{figure*}
    \centering
    \includegraphics[width=1.45\columnwidth]{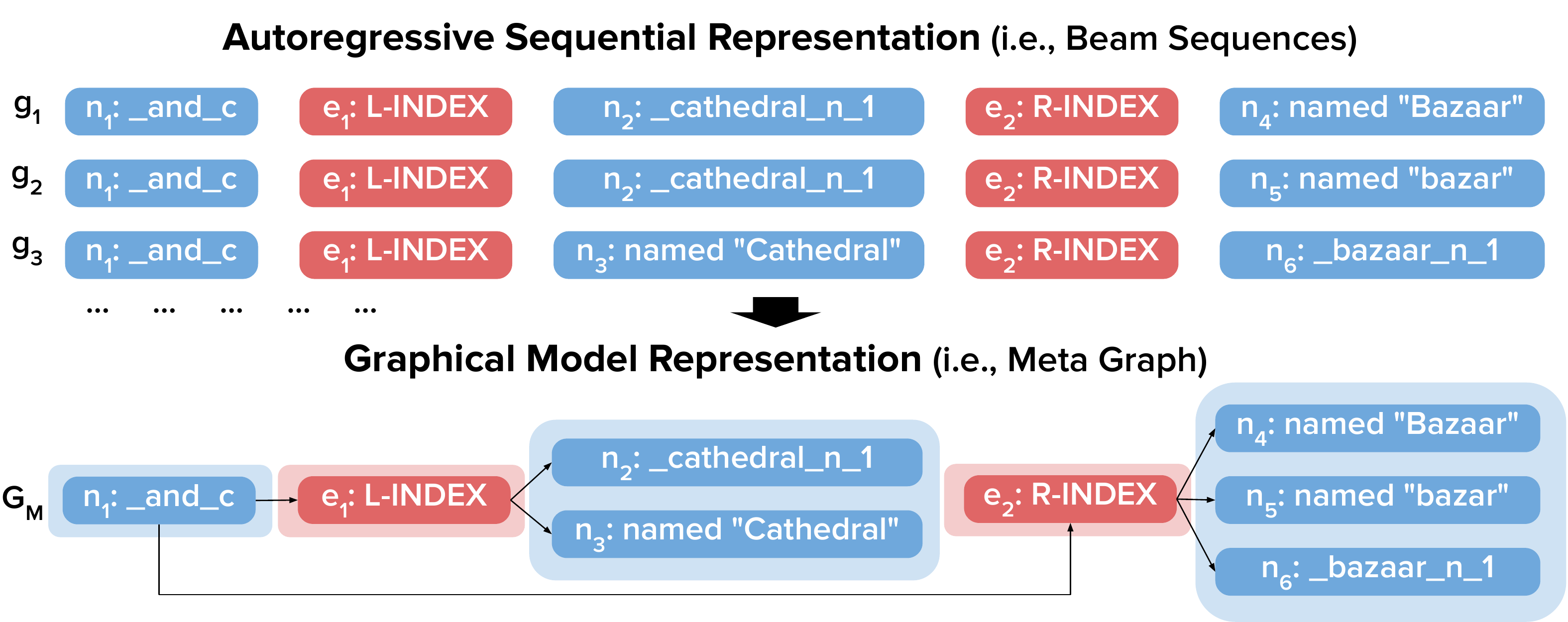}
    \caption{Visual illustration of constructing graphical model representation $\mathcal{G}_M$ from autorepressive representation $\{g_k\}_{k=1}^K$. The example here represents the sentence ``\textit{The Cathedral and the Bazaar}'' from the Eric Raymond Essay dataset. Note that here we have omitted the brackets in $g$ for simplicity (see \ref{fig:penman-exp}).
    }
    \label{fig:G-GAP}
\end{figure*}

%% file: algorithm/algorithm1.tex
\begin{algorithm}[ht]
\captionsetup{font=small}
\caption{Graph Autoregressive Process}\label{alg:GAP}
\small
\begin{minipage}{1.1\linewidth}
\begin{algorithmic}
\Inputs{Beam candidates with probabilities $\{p(g_k|x)\}_{k=1}^K$\\\textit{Meta graph} $\mathcal{G}$}
\Output{Marginal probabilities $\{p(s|x)\}$\\Graph model likelihood $\log p(G|x)$}
\For{$v \in G$}
\State Compute marginal likelihood:
\State $\qquad$ $p(v=s|x)$ (Equation \ref{eq:marginal-prob})
\State Compute graphical model likelihood:
\State $\qquad$ $\log p(v=s|\pa(v), x)$ (Equation \ref{eq:graphical-model-likelihood})
\EndFor
\Return{$\{p(v|x)\}, \log p(G|x))=\sum_{v\in G} \log p(v|\pa(v), x)$}
\end{algorithmic}
\end{minipage}
\end{algorithm}

%% file: sections/05.experiments.tex
\section{Experiments}
\label{sec:exp}
\subsection{Experiment Setup}
\smallsection{Datasets} Consistent with previous ERG works, we train the neural model on DeepBank v1.1 annotation of the Wall Stree Journal (WSJ), sections 00-21 (the same text annotated in the Penn Tree Bank) that correspond to ERG version 1214.

For OOD evaluation, 
we select 7 diverse datasets from the Redwoods Treebank corpus: Wikipedia (Wiki), the Brown Corpus (Brown), the Eric Raymond Essay (Essay), customer emails (E-commerce), meeting/hotel scheduling (Verbmobil), Norwegian tourism (LOGON) and the Tanaka Corpus (Tanaka) (See Appendix \ref{app:ood-datasets} for more details).

\smallsection{Model} Following \citet{lin-etal-2022-towards}, 
We train a T5$_{\texttt{large}}$ using the official T5X finetune pipeline\footnote{\url{https://github.com/google-research/t5x/blob/main/t5x/train.py}}, and use beam search with size $K=5$ at inference time. 
Further details are collected in Appendix \ref{app:hyperparameters}. 

\smallsection{Evaluation}
we use the standard eval metric \textsc{Smatch} \citep{cai-knight-2013-smatch}, which computes the maximum F1-score obtainable from an alignment between the predicted and gold graphs. We evaluate the models' average-case performance on all the 8 in-domain and OOD datasets, and also conduct fine-grained evaluation of the models' tail generalization performance across 19 important linguistic subcategories (Appendix \ref{app:fine-grained-phenomena}, Table \ref{tab:fine-grained}).

\input{sections/tables/ood-perf}
\smallsection{Baselines}
We compare with two recent state-of-the-art approaches from the neural-symbolic and ensemble graph parsing literature, respectively. (see Appendix \ref{sec:related-work} for a review) (1) \citet{lin-etal-2022-towards} is uncertainty-aware neural-symbolic framework method attained state-of-the-art performance on the in-domain DeepBank test set, and (2) \citet{hoang2021ensembling}, a majority-voting-based graph ensemble method that uses a voting strategy based on beam sequences from the T5 model and predictions from the ACE parser \footnote{We have tried several other variants for the voting candidates, e.g., top K predictions from the T5 parser and top 1 prediction + ACE prediction. It turns out the best one is using top K predictions from the T5 parser and ACE predictions.}. It doesn't exploit uncertainty. 

\subsection{Results}

The results are shown in Table \ref{tab:ood-results}. Detailed in-domain comparision with other previous work is in Appendix \ref{app:ind-test}. As shown, among the base models, the T5 and ACE parser achieve similar overall performance, with T5 strongly outperforms on in-domain data but underperforms on the OOD data (see last row in Table \ref{tab:ood-results}).
Our approach achieves best results on overall performance, which is $\sim35\%$ error reduction in aggregated \textsc{Smatch} score over the T5-based and symbolic approaches.

We now compare with the previous state-of-the-art methods. Though in-domain performance is not the focus of this work, our approach is still comparable to \textbf{Collab}, i.e., the neural-symbolic method from \citet{lin-etal-2022-towards}. However, on the challenging out-of-domain eval sets (e.g., E-commerce, Verbmobil whose topic and style are significantly different from WSJ), the performance of \textbf{Collab} starts to deteriorate. In comparison, our neural-symbolic approach remains robust out-of-domain. Its performance stays competitive with and even sometimes outperforms the ACE parser on difficult domains, illustrating the advantage of compositionality.


We also notice that the voting-based ensemble method \textbf{Vote} \citep{hoang2021ensembling} performs poorly in the neural-symbolic setting, despite based a moderate number of beam sequences. This is likely because the majority-voting approach requires a large number of diverse predictions from distinct models. When there are only two models, the ability of quantifying uncertainty becomes important.
\input{sections/tables/tail-perf}

\vspace{-0.5em}
\subsection{Fine-grained Linguistic Evaluation}
\label{sec:tail-perf}
ERG provides different levels of linguistic information that can benefit many NLP tasks, e.g., named entity recognition and semantic role labeling. This rich linguistic annotation provides an oppurtunity to evaluate model performance in meaningful population subgroups.
Detailed description of those linguistic phenomena is in Appendix \ref{app:fine-grained-phenomena}.

Result is in Table \ref{tab:fine-grained}. As shown, on OOD datasets, the T5 parser underperforms the ACE parser on most of the linguistic categories. Our approach outperforms both the neural model and the non-compositional neural-symbolic method especially on long-tail categories (the gray colored rows in the table), attaining an $>14\%$ average absolute gain compared to the base model. In some categories, our method even outperforms the ACE parser while all base model underperforms, e.g., \texttt{ARG3} of basic verb on Verbmobil and \texttt{ARG3} of verb-particle on E-commerce.

\subsection{Case Study: Synthesizing Novel Graphs}
\label{sec:case-study}
\input{sections/tables/novel-perf}
To test if our methods can generate optimal graph solution which the base models fail to obtain, we further explore the percentage of novel graphs (graphs that are not identical to any of the candidate predictions of the neural or symbolic model) for each dataset, and compare the corresponding \textsc{Smatch} scores on those novel cases. 
The results are shown in Table \ref{tab:novel-perf}. We see that our method synthesize novel graph parses that are in general of higher quality than that of the base models, thanks to the calibrated uncertainty (Section \ref{sec:compositional-uncertainty}). This indicates the compositional neural-symbolic inference can synthesize evidence across neural and symbolic results and produce novel graphs that are closer to ground truth.


%% file: sections/tables/ood-perf.tex
\begin{table}[ht]
\centering
\resizebox{\columnwidth}{!}{
\begin{tabular}{l|r|ccccc|cccc|c}
\toprule
   &  \textbf{\#}  & \textbf{T5} & \textbf{ACE} & \textbf{Vote} & \textbf{Collab.} & \textbf{Ours} &  \textbf{ACE*} \\
\midrule
WSJ (in-domain) & 1,437 & 96.56 & 87.14 & 88.22 & \textbf{97.01} & \underline{96.77} & 90.94\\
Wiki & 1,307 & 90.12 & 80.25 & 80.55 & \textbf{90.58} & \underline{90.04} & 90.42 \\
Brown & 2,182 & 92.05 & 91.74  & 85.46 & \textbf{93.58} & \underline{93.11} & 93.20 \\
Essay & 591 & 92.19 & 92.64 &  83.72 & \underline{93.57} & \textbf{93.76}  & 93.52 \\
E-commerce & 1,114 & 93.15 & \underline{97.25} & 87.38 & 95.44  & \textbf{97.37} & 98.36 \\
Verbmobil & 931 & 90.06 & \underline{95.15} & 84.80 & 92.24 & \textbf{96.42} & 97.62 \\
LOGON & 1,895 & 87.13 & \textbf{93.58} & 80.11 & 92.88 & \underline{93.33} & 94.17 \\
Tanaka & 2,796 & 95.24 & \textbf{98.38} & 91.03 & 96.79 & \underline{98.14} & 98.55 \\
\midrule
Mean w/ in-domain & - & 92.06 & 92.02 &  85.16 & \underline{94.01}  & \textbf{94.86} & 94.60 \\
Mean w/o in-domain & - & 91.50 & 92.63 & 84.72 & \underline{93.64} & \textbf{94.62} & 95.05\\
\bottomrule
\end{tabular}}
\caption{\textsc{Smatch} for T5, ACE, and collaborative/compostional inference. \# refers to the number of sentences in the dataset. ACE* refers to the evaluation results only for valid parse. Collab. refers to collaborative inference from \citet{lin-etal-2022-towards}. Vote refers to voting strategy from \citet{hoang2021ensembling}. The \textbf{bold} and \underline{underlined} refer to the best and the second best results.
}
\label{tab:ood-results}
\end{table}

%% file: sections/tables/tail-perf.tex
\begin{table*}[ht]
\centering
\resizebox{0.95\textwidth}{!}{
\begin{tabular}{l|rc|lll|rc|lll|rc|lll}
\toprule
& \multicolumn{5}{c|}{Essay} & \multicolumn{5}{c|}{E-commerce} & \multicolumn{5}{c}{Verbmobil} \\
Type & \# & ACE & \multicolumn{1}{c}{T5} & \multicolumn{1}{c}{Collab.} & Ours  & \# & ACE & \multicolumn{1}{c}{T5} & \multicolumn{1}{c}{Collab.} & Ours & \# & ACE & \multicolumn{1}{c}{T5} & \multicolumn{1}{c}{Collab.} & Ours\\\midrule
Compound           & 671 & 83.76 & 73.39 & 76.75 & \textbf{80.26} & 844   & 95.50 & 67.96 & 83.22 & \textbf{94.94} & 308 & 86.36 & 67.41 & 68.13 & \textbf{87.50*} \\
\rowcolor{gray!40}
\begin{tabular}[c]{@{}l@{}}Nominal {\it \scriptsize w/ nominalization}\end{tabular} & 15 & 80.00 & \textbf{80.00} &  73.33 & \textbf{80.00} & 6 & 100.00 & 77.78 & \textbf{100.00} & \textbf{100.00} & - & - & - & - & - \\
\begin{tabular}[c]{@{}l@{}}Nominal {\it \scriptsize w/ noun}\end{tabular} & 521  & 88.68 & 76.84 & 80.79 & \textbf{84.56} & 682 & 95.60 & 72.67 & 86.93 & \textbf{95.45} & 194 & 95.88 & 77.80 & 83.50 & \textbf{95.15} \\
\rowcolor{gray!40}
Verbal             & 18    & 72.22 & 57.89 & 73.68* & \textbf{78.95*}   & -     & -     & -     & -     & -     & - &  - & - & - & -\\
Named entity       & 74    & 67.57 & \textbf{71.05*} & 68.42* & 60.53   &  28   & 92.86 & 77.49 & 80.00 & \textbf{93.33*} & 80 & 62.50 & 56.51 & 52.50 & \textbf{67.50*}\\
\midrule
Argument structure & 3,314 & 87.09 & 82.63 & \textbf{85.52} & 85.26 & 5,932 & 95.79 & 83.60 & 88.47 & \textbf{94.73} & 4,206 & 95.29 & 77.52 & 86.57 & \textbf{94.56}\\
Total verb         & 1,616 & 83.66 & 81.11 & \textbf{83.78*} & 82.56 & 4,504 & 95.12 & 83.48 & 87.36 & \textbf{93.90} & 2,330 & 95.19 & 82.25 & 89.36 & \textbf{94.35}\\
Basic verb         & 895   & 83.35 & 82.01 & \textbf{84.71*} & 83.50* & 2,910 & 94.85 & 87.20 & 90.14 & \textbf{92.77} & 1,206 & 94.36 & 89.15 & 91.48 & \textbf{94.48*}\\
\ \ \textit{ARG1}  & 694   & 88.90 & 88.26 & \textbf{90.61*} & 88.62 & 2,494 & 96.79 & 95.64 & 97.08* & \textbf{97.31*} & 1,168 & 96.75 & 95.40 & 95.27 & \textbf{96.90*}\\
\ \ \textit{ARG2}  & 708   & 88.28 & 86.69 & \textbf{89.04*} & 88.77* & 2,660 & 97.14 & 91.11 & 93.36 & \textbf{97.20*} & 876   & 95.89 & 89.34 & 93.91 & \textbf{95.65}\\
\rowcolor{gray!40}
\ \ \textit{ARG3}  & 69    & 83.61 & 78.57 & 78.57 & \textbf{80.14} & 382   & 90.05 & 70.91 & 75.13 & \textbf{78.07} & 62    & 93.55 & 67.56 & 87.50 & \textbf{96.88*}\\
    Verb-particle  & 721   & 84.05 & 79.99 & 65.15 & \textbf{81.41} & 1,592 & 95.61 & 76.94 & 82.31 & \textbf{95.95*} & 1,124 & 96.09 & 74.14 & 87.07 & \textbf{94.22}\\
\ \ \textit{ARG1}  & 620   & 87.90 & 84.39 & \textbf{86.53} & 85.58 & 1,448 & 96.27 & 80.77 & 84.73 & \textbf{96.62*} & 1,096 & 96.53 & 80.20 & 90.77 & \textbf{96.90*}\\
\ \ \textit{ARG2}  & 498   & 86.14 & 84.96 & 86.52* & \textbf{88.77*} & 888   & 96.85 & 71.30 & 81.33 & \textbf{95.56} & 424  & 94.34 & 66.73 & 78.90 & \textbf{92.66}\\
\rowcolor{gray!40}
\ \ \textit{ARG3}  & 62    & 79.03 & 65.15 & 65.15 & \textbf{74.24} & 208   & 93.27 & 69.05 & 83.02 & \textbf{96.23*} & 24    & 83.33 & 47.17 & \textbf{58.33} & \textbf{58.33}\\
\rowcolor{gray!40}
Total noun         & 189   & 91.53 & 82.90 & 86.01 & \textbf{86.49} & 90    & 100.00 & 76.81 & 78.26 & \textbf{97.83} & 26    & 92.31 & 69.00 & \textbf{93.33*} & \textbf{93.33*}\\
Total adjective    & 1,336 & 90.64 & 84.36 & 87.13 & \textbf{88.39} & 1,116 & 97.67 & 84.62 & 93.07 & \textbf{97.34} & 1,838 & 95.43 & 72.54 & 82.75 & \textbf{94.81}\\
\midrule
Reentrancy         & 850   & 80.59 & 78.39 & \textbf{81.26*} & 77.01 & 1,686 & 95.73 & 75.83 & 81.59 & \textbf{84.76} & 800   & 93.25 & 60.23 & 72.77 & \textbf{89.20}\\
\ \ \textit{passive} & 173 & 86.71 & 83.33 & \textbf{88.89*} & 86.71 & 222   & 98.20 & 85.56 & 92.11 & \textbf{97.37} & 12    & 100.00 & 79.10 & \textbf{100.00} & \textbf{100.00}\\
\bottomrule
\end{tabular}}
\caption{Comparing ACE, Collab. \citep{lin-etal-2022-towards} and our parsers on fine-grained linguistic categories. All scores are reported in accuracy. The \colorbox{gray!40}{gray colored row} means long-tail phenomenon ($<500$ cases in the training set). The \textbf{bold} indicates the best results among neural approaches (T5, Collab. and Ours). * indicates the result is better than ACE parser.
}
\label{tab:fine-grained}
\end{table*}

%% file: sections/tables/novel-perf.tex
\begin{table*}[ht]
\centering
\resizebox{1.4\columnwidth}{!}{
\begin{tabular}{l|r|cccccccc}
\toprule
 & \textbf{\%} & \textbf{Top 1} & \textbf{Top 2} & \textbf{Top 3} & \textbf{Top 4} & \textbf{Top 5} & \textbf{Collab.} & \textbf{ACE} & \textbf{Ours} \\
 \midrule
In-domain  & 31.25 & 94.95 & 93.01 & 91.91 & 89.92 & 89.58 & 95.10 & 82.80 & \textbf{98.44} \\
Wiki       & 32.29 & 87.55 & 86.54 & 85.56 & 86.00 & 83.90 & 88.77 &  82.67 &\textbf{92.24} \\
Brown      & 46.84 & 90.54 & 89.34 & 88.57 & 88.10 & 87.11 & 92.53 & 96.15 &\textbf{96.56} \\
Essay      & 50.93 & 90.71 & 90.02 & 89.31 & 89.02 & 87.60 & 92.41 & 95.73 &\textbf{96.08} \\
E-commerce & 34.65 & 90.03 & 88.34 & 86.61 & 85.56 & 82.91 & 92.82 & \textbf{98.96} & 97.54 \\
Verbmobil  & 39.96 & 85.45 & 83.06 & 81.54 & 79.30 & 78.27 & 88.42 & \textbf{97.78} & 96.70 \\
LOGON      & 58.10 & 90.75 & 89.65 & 88.20 & 87.90 & 86.95 & 92.50 &  96.70 & \textbf{97.06} \\
Tanaka     & 24.89 & 89.35 & 87.46 & 85.60 & 83.55 & 83.16 & 92.30 & 98.23 & \textbf{98.27} \\
\midrule
All        & 38.76 & 90.57 & 89.18 & 88.01 & 87.24 & 86.13 & 92.29 & 93.93 & \textbf{96.28} \\
\bottomrule
\end{tabular}}
\caption{\textsc{Smatch} performance on novel graphs, where the results of our inference process are not identical to any of the candidates from the base model.}
\label{tab:novel-perf}
\end{table*}

%% file: sections/06.related-work.tex
\section{Related Work}
\label{sec:related-work}
In this section we introduce related work for neural-symbolic and ensemble learning for graph semantic parsing. For a broader context of graph semantic parsing, please refer to Appendix~\ref{app:gsp-review}.

\smallsection{Neural-Symbolic Graph Semantic Parsing}
Though neural models excel at semantic parsing, they have been shown to struggle with out-of-distribution compositional generalization, while grammar or rule-based approaches work relatively robustly. This has motivated the work in neural-symbolic parsing where symbolic approaches are imported as inductive bias \citep{shaw-etal-2021-compositional, kim2021sequence, cheng-etal-2019-learning, cole-etal-2021-graph-based}.
For graph meaning representations, importing inductive bias into neural model was somehow difficult due to the much  more complicated structure compared to pure syntactic rules or logical formalism \citep{peng-etal-2015-synchronous,peng-gildea-2016-uofr}. To address this, \citet{lin-etal-2022-towards} proposes a collaborative framework by designing a decision criterion for beam search that incorporates the prior knowledge from a symbolic parser and accounts for model uncertainty, which achieves the state-of-the-art results on the in-domain test set.

\smallsection{Ensemble Learning for Graph Parsing}
Ensemble learning is a popular machine learning approach that combines predictions from multiple candidates to create a new one that is more robust and accurate than individual predictions.
Previous studies have explored various ensemble learning approaches for graph parsing \citep{green-zabokrtsky-2012-hybrid,barzdins-gosko-2016-riga}.
Specifically, for graph semantic parsing at subgraph level,
\citet{hoang2021ensembling} make use of checkpoints from models of different architectures, and mining the largest graph that is the most supported by a collection of graph predictions. They then propose a heuristic algorithm to approximate the optimal solution.

Compare to the previous ensemble work, our work differ in three ways: (1) Our decision rule is based on neural model confidence, so the decision is driven not by model consensus, but by model confidence which indicates when the main (neural) result is untrustworthy and needs to be complemented by symbolic result. Model consensus is effective when there exists a large number of candidate models. However, in the neural-symbolic setting when there are only two models, the ability of quantifying model uncertainty becomes important. (2) A secondary contribution of our work is to produce an parsing approach for the ERG community that not only exhibits strong average-case performance on in-domain and OOD environments, but also generalizes robustly in important categories of tail linguistic phenomena. Therefore, our investigation goes beyond average-case performance and evaluates in tail generalization as well. (3) We reveal a more nuance picture of neural models' OOD performance: a neural model's top K parses in fact often contains subgraphs that generalize well to OOD scenarios, but the vanilla MLE-based inference fails to select them (see Section \ref{sec:case-study} for more details).

%% file: sections/07.conclusion.tex
\section{Conclusions}
We have shown how to perform accurate and robust semantic parsing across a diverse range of genres and linguistic categories for English Resource Grammar. We achieve this by taking the advantage of both the symbolic parser (ACE) and the neural parser (T5) at a fine-grained subgraph level using compositional uncertainty, an aspect missing in the previous neural-symbolic or ensemble parsing work. 
Our approach attains the best known result on the aggregated SMATCH score across eight evaluation corpus from Redwoods Treebank, attaining $35.26 \%$ and $35.60 \%$ error reduction
over the neural and symbolic parser, respectively.

%% file: sections/appendices/meaning-representation.tex
\section{Graph-based Meaning Representation}
\label{app:meaning-representation}
Considerable NLP research has been devoted to the transformation of natural language utterances into a desired linguistically motivated semantic representation. Such a representation can be understood as a class of discrete structures that describe lexical, syntactic, semantic, pragmatic, as well as many other aspects of the phenomenon of human language.
In this domain, graph-based representations provide a light-weight yet effective way to encode rich semantic information of natural language sentences and have been receiving heightened attention in recent years. Popular frameworks under this umbrella includes Bi-lexical Semantic Dependency Graphs \citep[SDG;][]{bos2004wide,ivanova-etal-2012-contrastive,oepen-etal-2015-semeval}, Abstract Meaning Representation \citep[AMR;][]{banarescu-etal-2013-abstract}, Graph-based Representations for English Resource Grammar \citep[ERG;][]{oepen-lonning-2006-discriminant,copestake-2009-invited}, and Universal Conceptual Cognitive Annotation \citep[UCCA;][]{abend-rappoport-2013-universal}.

%% file: sections/appendices/graph-semantic-parsing.tex
\section{Literature Review on Graph Semantic Parsing}
\label{app:gsp-review}
In this section, we present a summary of different parsing technologies for graph-based meaning representations in addition to the ones discussed in \ref{sec:parsing-to-erg}, with a focus on English Resource Grammar (ERG).

\paragraph{Grammar-based approach}
In this type of approach, a semantic graph is derived according to a set of lexical and syntactico-semantic rules. For ERG parsing, sentences are parsed to HPSG derivations consistent with ERG. The nodes in the derivation trees are feature structures, from which MRS is extracted through unification.
The parser has a default parse ranking procedure trained on a treebank, where maximum entropy models are used to score the derivations in order to find the most likely parse. However, this approach fails to parse sentences for which no valid derivation is found \citep{toutanova2005stochastic}.
There are two main existing grammar-based parsers for ERG parsing: the PET system \citep{callmeier2000pet} and the ACE system \citep{crysmann-packard-2012-towards}. The core algorithms implemented by both systems are the same, but ACE is faster in certain common configurations. We choose ACE as the symbolic parser in our work.

\paragraph{Factorization-based approach}
This type of approach is inspired by graph-based dependency tree parsing \citep{mcdonald2006discriminative}. A factorization-based parser explicitly models the target semantic structures by defining a score function that can evaluate the probability of any candidate graph. For ERG parsing, \citet{cao-etal-2021-comparing} implemented a two-step pipeline architecture that identifies the concept nodes and dependencies by solving two optimization problems, where prediction of the first step is utilized as the input for the second step. \citet{chen-etal-2019-peking} presented a four-stage pipeline to incrementally construct an ERG graph, whose core idea is similar to previous work.

\paragraph{Transition-based approach}
In these parsing systems, the meaning representations graph is generated via a series of actions, in a process that is very similar to dependency tree parsing \cite{yamada-matsumoto-2003-statistical,nivre-2008-algorithms}, with the difference being that the actions for graph parsing need to allow reentrancies. For ERG parsing,
\citet{buys-blunsom-2017-robust} proposed a neural encoder-decoder transition-based parser, which uses stack-based embedding features to predict graphs jointly with unlexicalized predicates and their token alignments.

\paragraph{Composition-based approach}
Following a principle of compositionality, a semantic graph can be viewed as the result of a derivation process, in which a set of lexical and syntactico-semantic rules are iteratively applied and evaluated. For ERG parsing, based on \citet{chen-etal-2018-accurate}, \citet{chen-etal-2019-peking} proposed a composition-based parser whose core engine is a graph rewriting system that explicitly explores the syntactico-semantic recursive derivations that are governed by a synchronous SHRG.

\paragraph{Translation-based approach}
This type of approach is inspired by the success of seq2seq models which are the heart of modern Neural Machine Translation. A translation-based parser encodes and views a target semantic graph as a string from another language.
In a broader context of graph semantic parsing, simply applying seq2seq models is not successful, in part because effective linearization (encoding graphs as linear sequences) and data sparsity were thought to pose significant challenges \citep{konstas-etal-2017-neural}.
Alternatively, some specifically designed preprocessing procedures for vocabulary and entities can help to address these issues \citep{konstas-etal-2017-neural, peng-etal-2017-addressing}.
These preprocessing procedures are very specific to a certain type of meaning representation and are difficult to transfer to others.
To address this, \citet{lin-etal-2022-towards} propose a variable-free top-down linearization and a compositionality-aware tokenization for ERG graph preprocessing, and successfully transfer the ERG parsing into a translation problem that can be solved by a state-of-the-art seq2seq model T5 \citep{raffel2020exploring}. The parser achieves the best known results on the in-domain test set from the DeepBank benchmark.

%% file: sections/appendices/additional-discussion.tex
\section{Additional Methods Discussions}
\label{app:additional-discuss}

\subsection{Efficient Probability Estimation Using Beam Outputs}
\label{app:prob-estimation}

The marginalized probability $\hat{p}(s_i|x)$ provides a way to reason about the \textit{global} importance of $s_i$ by integrating the probabilistic  evidence $p(s_i|s_{k, <i}, x)$ over the whole beam-sampled posterior space. It is able to capture the cases of spurious graph elements $s_i$ with high local probability $p(s_i|s_{k, <i}, x)$ but low global likelihood (i.e., only appear in a few low-probability beam candidates), which is useful for inferring sparse global structures for the meta graph (Appendix \ref{app:extension}).

In the importance weight $\pi_k$, the temperature parameter $t$ controls how evidence for $p(s_i|x)$ is aggregated across beam samples $\{g_k\}_{k=1}^K$. When $t\rightarrow 0$, the above is equivalent to selecting $p(s_i|s_{k, <i}, x)$ from the most probable subsequence $s_{k, <i}$; when $t \rightarrow \infty$, the above is equivalent to simple averaging of $p(s_i|s_{k, <i}, x)$ from all beam candidates. In the experiments, we find that the value of $t$ does not have a significant impact on the final performance. In general, we recommend fixing it to a small value (e.g., $t=0.1$) to suitably downweighting the contribution from improbable beam candidates.

\section{Simplified Expression for Graphical Model Likelihood}
\label{app:likelihood-computation}

Given the candidates graphs $\{G_k\}_{k=1}^K$, we can express the likelihood for $p(v|\pa(v), x)$ by writing down a multinomial likelihood enumerating over different values of $\pa(v)$ \citep{murphy2012machine}.
For example, say $\pa(n)=(e_1, e_2)$ which represents a subgraph of two edges $(e_1, e_2)$ pointing into a node $n$. Then the conditional probability $p(n|\pa(n), x)$ can be computed by enumerating over the observed values of $(e_1, e_2)$ pair:
\resizebox{\linewidth}{!}{
\begin{minipage}{\linewidth}
\begin{align*}
    & p(n|\pa(n), x) = p(n|(e_1, e_2), x) \\
    &\propto 
    \prod_{c\in \text{Candidate}(e_1, e_2)} 
    p(n|(e_1, e_2)=c, x)^{K_c}
\end{align*}
\end{minipage}
}
where $\text{Candidate}(e)$ is the collection of possible symbols $s$ the variable $e$ can take, and $K_c$ is the number of times $(e_1, e_2)$ takes a particular value $c\in \text{Candidate}(e_1, e_2)=\text{Candidate}(e_1) \times \text{Candidate}(e_2)$. 

Then, the log likelihood becomes:
\begin{align*}
    &\log p(n|\pa(n), x) \\
    &= \sum_{c} K_c*\log p (n|(e_1, e_2)=c)
\end{align*}
To simplify this above expression, we notice that $\log p(n|\pa(n), x)$ can be divided by the constant beam size K without impacting the inference. As a result, the log probability can be computed by simplify averaging the values of $\log p(v|\pa(v)=c_k)$ across the beam candidates:
\begin{align*}
    & \log p(n|\pa(n), x) \\
    &\propto \sum_{c} \frac{K_c}{K} \log p(n|(e_1, e_2)=c) \\
    &= \frac{1}{K} \sum_{k=1}^K \log p(n|(e_1, e_2)=c_k)
\end{align*}
where $c_k$ is the value of $(e_1, e_2)$ in $k^\text{th}$ beam candidate.

%% file: sections/appendices/extension-implementation.tex
\section{Extensions and Practical Implementation}
\label{app:extension}
\subsection{Infer Sparse Global Structure via Likelihood-based Pruning}
In practice, the meta graph$\mathcal{G}$ can contain spurious elements $v$ that have a high local likelihoods $\log p(v|\pa(v), x)$ but very low global probabilities $p(v|x)$. This happens when the element $v$ only appears in a few low-probability beam sequences. These spurious nodes and edges often adds redundancy to the generated graph (i.e., hurting precision), and cannot be eliminated by the neural-symbolic inference procedure, due to their high local conditional probability $p(v|\pa(v), x)$.

Consequently, we find it empirically effective to perform sparse structure inference for$\mathcal{G}$ based on global probabilities $p(v|x)$ before diving into local neural-symbolic prediction for graph components. In this work, we carry out this global structure inference by considering a simple threshold-and-project procedure, i.e., pruning out all the graph elements whose global probability $||p(v|x)||_\infty = \max_{s\in\text{Candidate}(v)}p(v=s|x)$ is lower than a threshold $t$, but will keep $v$ if its removal will lead to an invalid graph with disconnected subcomponents. Here $||p(v|x)||_\infty$ is the total variation metric that returns the maximum probability. 

Algorithm \ref{alg:pruning} summarizes this procedure. From a theoretical perspective, this is equivalent to finding the most sparse solution with respect to threshold $t$ within the space of valid (i.e., connected) subgraphs of$\mathcal{G}$.

\input{algorithm/algorithm3}

\begin{figure}[ht]
    \centering
    \includegraphics[width=\columnwidth]{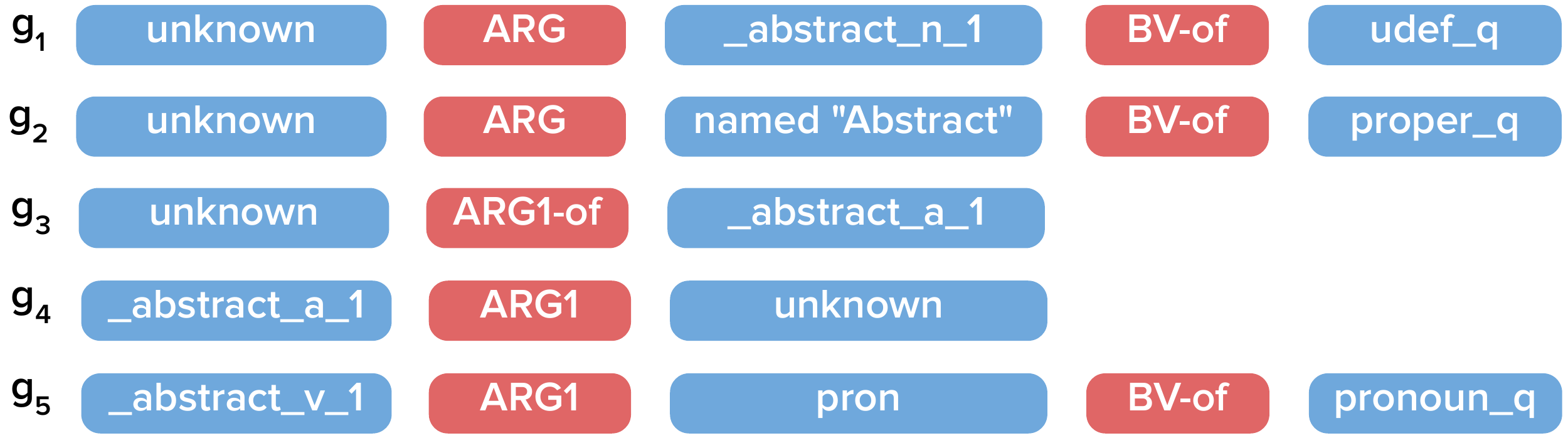}
    \caption{Autoregressive Representation (i.e., beam sequences) for the sentence ``\textit{Abstract}'' from the Eric Raymond Essay dataset. Note that $g_3$ and $g_4$ are actually the same graph but with different linearization orders.}
    \label{fig:cluster-exp}
\end{figure}

\subsection{Handle Multi-modality via Mixture Modeling}
In some rare cases where the input sentence is fragmented or ill-formed, the neural model may output multiple beam sequences with drastically different high-level structures, creating difficulty for the graph merging procedure (See Figure \ref{fig:cluster-exp} for an example).

We can handle this multi-modality in observed graph structure by extending $p(G|x)$ to be a mixture of GAP distributions, so that the graphical model likelihood becomes:
\begin{align*}
    p(G|x) = \sum_{m\in M} p(G|m, x)p(m|x)
\end{align*}
where $p(m|x)$ is a categorical distribution over the mixture components $m \in M$. Here each component $m$ induce a meta graph $\mathcal{G}_m$ for graph $G_m=\langle \mathbf{N}_m, \mathbf{E}_m \rangle$, such that
\begin{align*}
    &p(G|m, x) = p(G_m|x) = \prod_{v\in G_m} p(v|\pa(v), x)\\
    &= \prod_{n\in \mathbf{N}_m} p(n|\pa(n), x) * \prod_{e\in \mathbf{E}_m} p(e|\pa(e), x)
\end{align*}
Given beam sequences $\{g_k\}_{k=1}^K$, the mixture components can be estimated using a standard clustering algorithm based on an edit distance between beam candidate $g_k$. Based on our experiments, hierarchical agglomerative clustering (HAC) combined with the longest common subsequence (LCS) distance often leads to the best result. After clustering, $p(m|x)$ is computed as the empirical probability of beam sequences belonging the $m^\text{th}$ cluster, and the meta graph $\mathcal{G}_m$ is computed by applying the graph merging procedure to the beam sequences in the $m^\text{th}$ cluster.

To conduct neural symbolic inference, we also need to define the symbolic prior $p_0$ for the mixture distribution:
\begin{align*}
    p_0(G) &= \sum_{m \in M} p_0(G|m) * p_0(m) \\
           &= \sum_{m \in M} [\prod_{v\in G_m} p_0(v)* p_0(m)]
\end{align*}
where $p_0(v=s) \propto \exp(I(s\in G_0))$ as define previously, and we define $p_0(m)=\exp(-\text{\textsc{Smatch}}(G_m, G_0))$ following the previous work \citep{lin-etal-2022-towards}.

As a result, the decision criteria for neural-symbolic inference under the mixture model becomes:
\begin{align*}
    R(G_m|x)=R(m|x)+\sum_{v\in G_m}R(v|x)
\end{align*}
where $\sum_{v\in G_m}R(v|x)$ is the component-wise decision criteria as defined in the main text, and $R(m|x)$ is the additional term for the mixture components:
\begin{align*}
    R(m|x) &= \alpha(m|x) * \log p(m|x) \\
    &+ (1-\alpha(m|x)) * \log p_0(m)
\end{align*}
where $\alpha(m|x)=\sigma(-\frac{1}{T}H(m|x)+b)$ is the trade-off parameter driven by the average log likelihood of beam sequences in the $m^\text{th}$ cluster $C_m$, i.e., $H(m|x)=\frac{1}{|C_m|}\sum_{g_k \in C_m}-\log (g_k|x)$.

During inference, we can again proceed in a greedy fashion, first select the optimal $\hat{m}$ based on $R(m|x)$, and then perform compositional neural-symbolic inference with respect to $\mathcal{G}_{\hat{m}}$ using $\sum_{v\in G_{\hat{m}}}R(v|x)$.

\input{algorithm/algorithm4}

As a result, the complete precedure with all optional extensions are shown in Algorithm \ref{alg:complete-procedure}.

%% file: algorithm/algorithm3.tex
\begin{algorithm}[ht]
\captionsetup{font=small}
\caption{Likelihood-based Pruning}\label{alg:pruning}
\small
\begin{minipage}{1.1\linewidth}
\begin{algorithmic}
\Inputs{Meta Graph $\mathcal{G}_M$\\Marginal probabilities $\{p(s|x)\}_{s\in G}$\\Threshold $t$}
\Output{Pruned graph $\mathcal{G}_M'$}
\Initialize{$\mathcal{G}'_M =\mathcal{G}_M$}
\For{$v \in \mathcal{G}'_M$}
\If{$||p(v|x)||_\infty < t$ AND $\mathcal{G}'_M  \setminus\{v\}$ is connected}
\State Prune $v:\mathcal{G}'_M =\mathcal{G}'_M  \setminus\{v\}$
\EndIf
\EndFor
\Return{$\mathcal{G}_M'$}
\end{algorithmic}
\end{minipage}
\end{algorithm}

%% file: algorithm/algorithm4.tex
\begin{algorithm}[ht]
\captionsetup{font=small}
\caption{Complete Procedure with All Extensions}\label{alg:complete-procedure}
\small
\begin{minipage}{1.1\linewidth}
\begin{algorithmic}
\Inputs{Beam candidates and associated token-level probabilities $\{p(g_k|x)\}_{k=1}^K$}
\Output{Neural-symbolic graph prediction $G$}
\OpEstimate{Mixture components $\{\mathcal{G}_m\}_{m=1}^M$, $\{p(m|x)_{m=1}^M\}$ from \textbf{Cluster} $\{p(g_k|x)\}_{k=1}^K$\\
\vspace{0.8em}
Optimal mixture components $\mathcal{G}=\mathcal{G}_{\hat{m}}$, where $\hat{m}=\argmax R(m|x)$}
\Estimate{Marginal probability and graphical model likelihood (Algorithm \ref{alg:GAP}):
\vspace{-0.5em}
\begin{align*}
    \{p(v|x)\}_{v\in G}, \log p(G|x) = \text{\textbf{GAP}}(\mathcal{G})
\end{align*}}
\vspace{-1.2em}
\Infer{Global graph structure via likelihood-based pruning (Algorithm \ref{alg:pruning})
\vspace{-0.5em}
\begin{align*}
    G'=\text{\textbf{ThresholdAndProject}}(G, \{p(v|x)\}_{v\in G})
\end{align*}
Local node / edge prediction via compostitional neural-symbolic inference (Algorithm \ref{alg:CNSI})
\vspace{-0.5em}
\begin{align*}
    G = \text{\textbf{NeuralSymbolicInference}(G')}
\end{align*}}
\end{algorithmic}
\end{minipage}
\end{algorithm}

%% file: sections/appendices/graph-alignment-algorithm.tex
\section{Graph Matching Algorithm}
\label{app:graph-align}
In general, finding the largest common subgraph is a well-known
computationally intractable problem in graph theory. However, for graph parsing problems where graphs have labels and a simple tree-like structure, some efficient heuristics are proposed to approximate the best match by a hill-climbing algorithm \citep{cai-knight-2013-smatch}. The initial match is modified iteratively to optimize the total number of matches with a predefined number of iterations (default value set to $5$). This algorithm is very efficient and effective, it was also used to calculate the \textsc{Smatch} score in \citet{cai-knight-2013-smatch}.

%% file: sections/appendices/ood-datasets.tex
\section{Details for OOD Datasets}
\label{app:ood-datasets}

\paragraph{Wikipedia (Wiki)} The DeepBank team constructed a treebank for 100 Wikipedia articles on Computational Linguistics and closely related topics. The treebank of 11,558 sentences comprises 16 sets of articles. The corpus contains mostly declarative, relatively long sentences, along with some fragments.

\paragraph{The Brown Corpus (Brown)} The Brown Corpus was a carefully compiled selection of current American English, totalling about a million words drawn from a wide variety of sources.

\paragraph{The Eric Raymond Essay (Essay)} The treebank is based on translations of the essay ``The Cathedral and the Bazaar'' by Eric Raymond. The average length and the linguistic complexity of these sentences is markedly higher than the other treebanked corpora.

\paragraph{E-commerce} While the ERG was being used in a commercial software product developed by the YY Software Corporation for automated response to customer emails, a corpus of training and test data was constructed and made freely available, consisting of email messages composed by people pretending to be customers of a fictional consumer products online store. The messages in the corpus fall into four roughly equal-sized categories: Product Availability, Order Status, Order Cancellation, and Product Return.

\paragraph{Meeting/hotel scheduling (Verbmobil)} This dataset is a collection of transcriptions of spoken dialogues, each of which reflected a negotiation either to schedule a meeting, or to plan a hotel stay. One dialogue usually consists of 20-30 turns, with most of the utterances relatively short, including greetings and closings, and not surprisingly with a high frequency of time and date expressions as well as questions and sentence fragments.

\paragraph{Norwegian tourism (LOGON)} The Norwegian/English machine translation research project LOGON acquired for its development and evaluation corpus a set of tourism brochures originally written in Norwegian and then professionally translated into English. The corpus consists almost entirely of declarative sentences and many sentence fragments, where the average number of tokens per item is higher than in the Verbmobil and E-commerce data.

\paragraph{The Tanaka Corpus (Tanaka)} This treebank is based on parallel Japanese-English sentences, which was adopted to be used with in the WWWJDIC dictionary server as a set of example sentences associated within words in the dictionary.

%% file: sections/appendices/implementation.tex
\section{Implementation and Hyperparameters}
\label{app:hyperparameters}
\paragraph{T5 Model} We use the open-sourced T5X \footnote{\url{https://github.com/google-research/t5x}}, which is a new and improved implementation of T5 codebase in JAX and Flax.
Specifically, we use the official pretrained T5-Large (770 million parameters), which is the same size as the one used in \citet{lin-etal-2022-towards}, and finetuned it on DeepBank in-domain training set. Specifically, the total training step is 1,750,000 including 1,000,000 pretrain steps. For fine-tuning the T5 model on ERG parsing, batch size is set to 128, the output and input sequence length is set to $512$, and dropout rate is set to $0.1$.

\paragraph{Hyperparameters} For the trade-off parameter $\alpha(v|x)=\sigma(-\frac{1}{T}H(v|x)+b)$, we set temperature $T=0.1$ and bias $b=0.25$.

%% file: sections/appendices/ind-test.tex
\begin{figure*}[ht]
 \centering
 \subfigure{
    \includegraphics[width=0.4\linewidth]{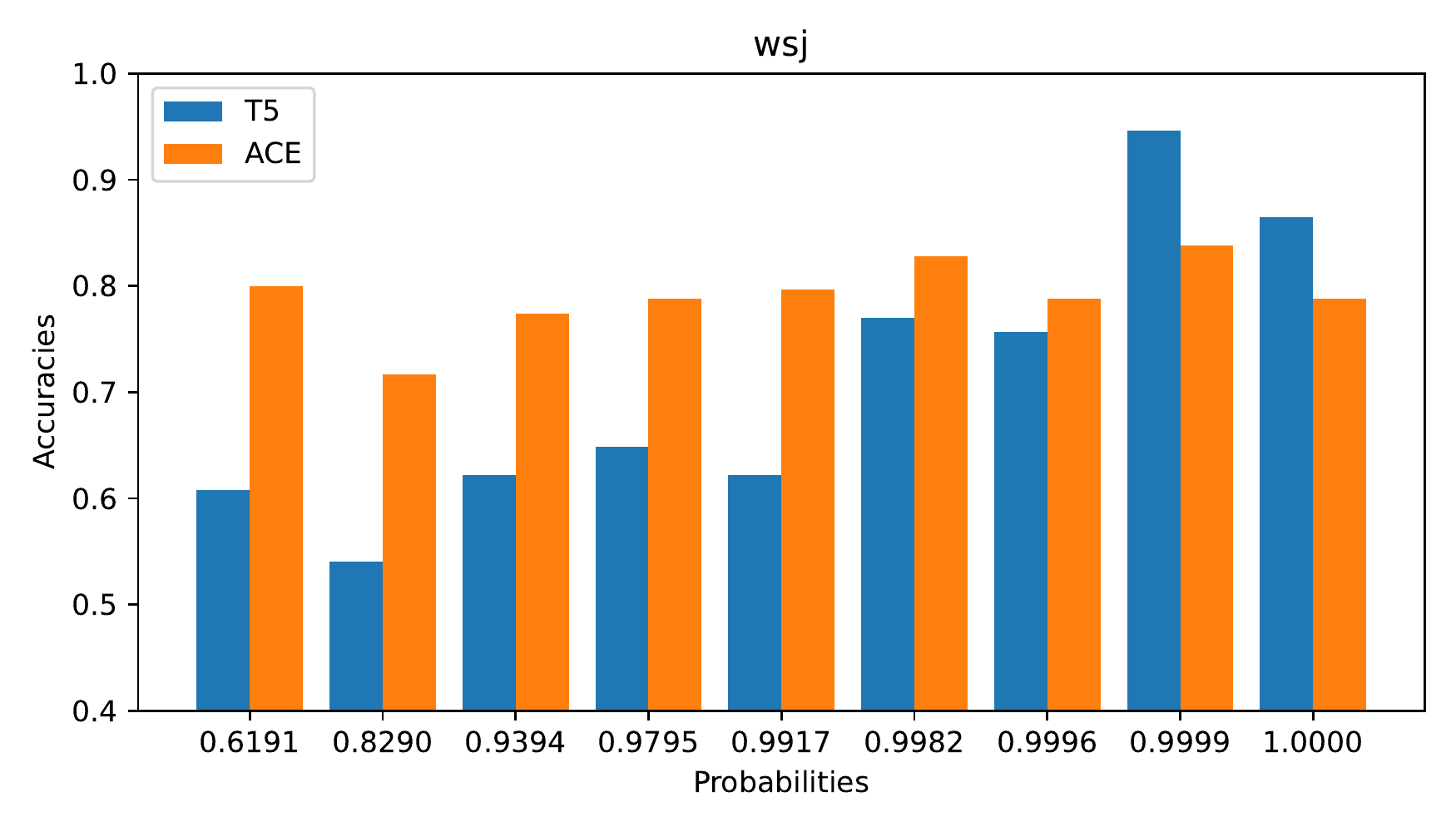}
 }
 \subfigure{
    \includegraphics[width=0.4\linewidth]{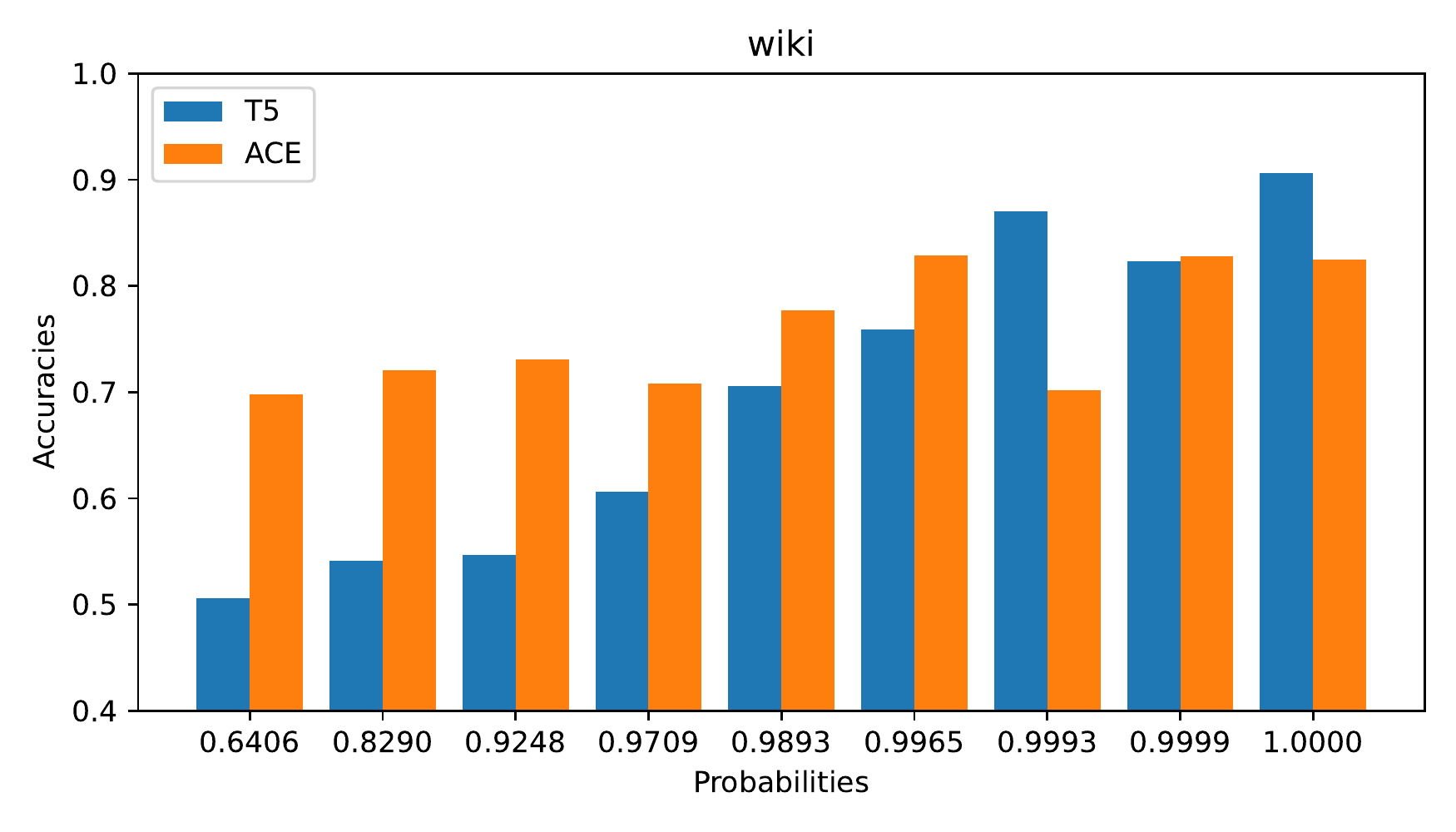} 
 }\\\vspace{-1.2em}
  \subfigure{
    \includegraphics[width=0.4\linewidth]{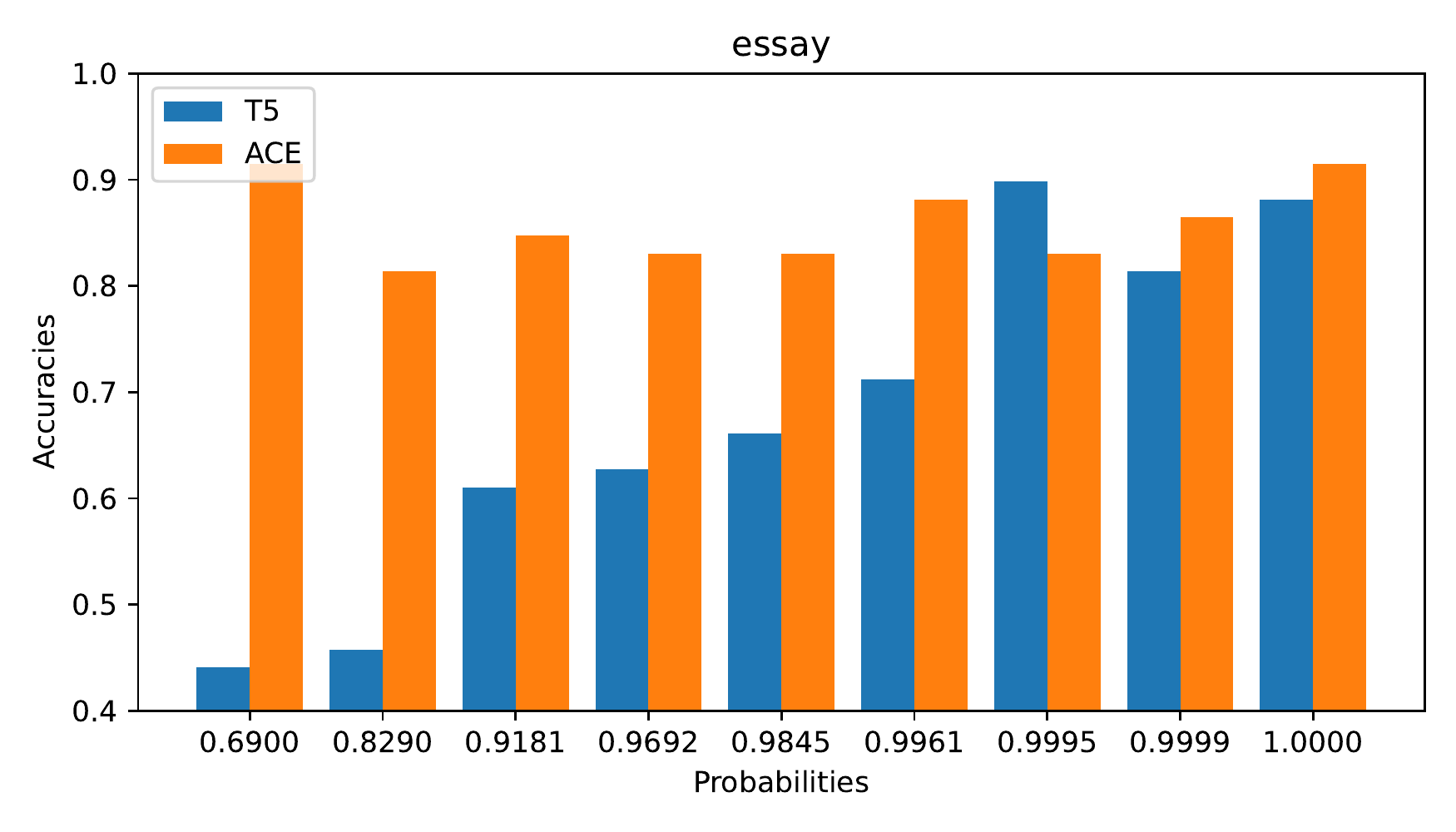} 
 }
  \subfigure{
    \includegraphics[width=0.4\linewidth]{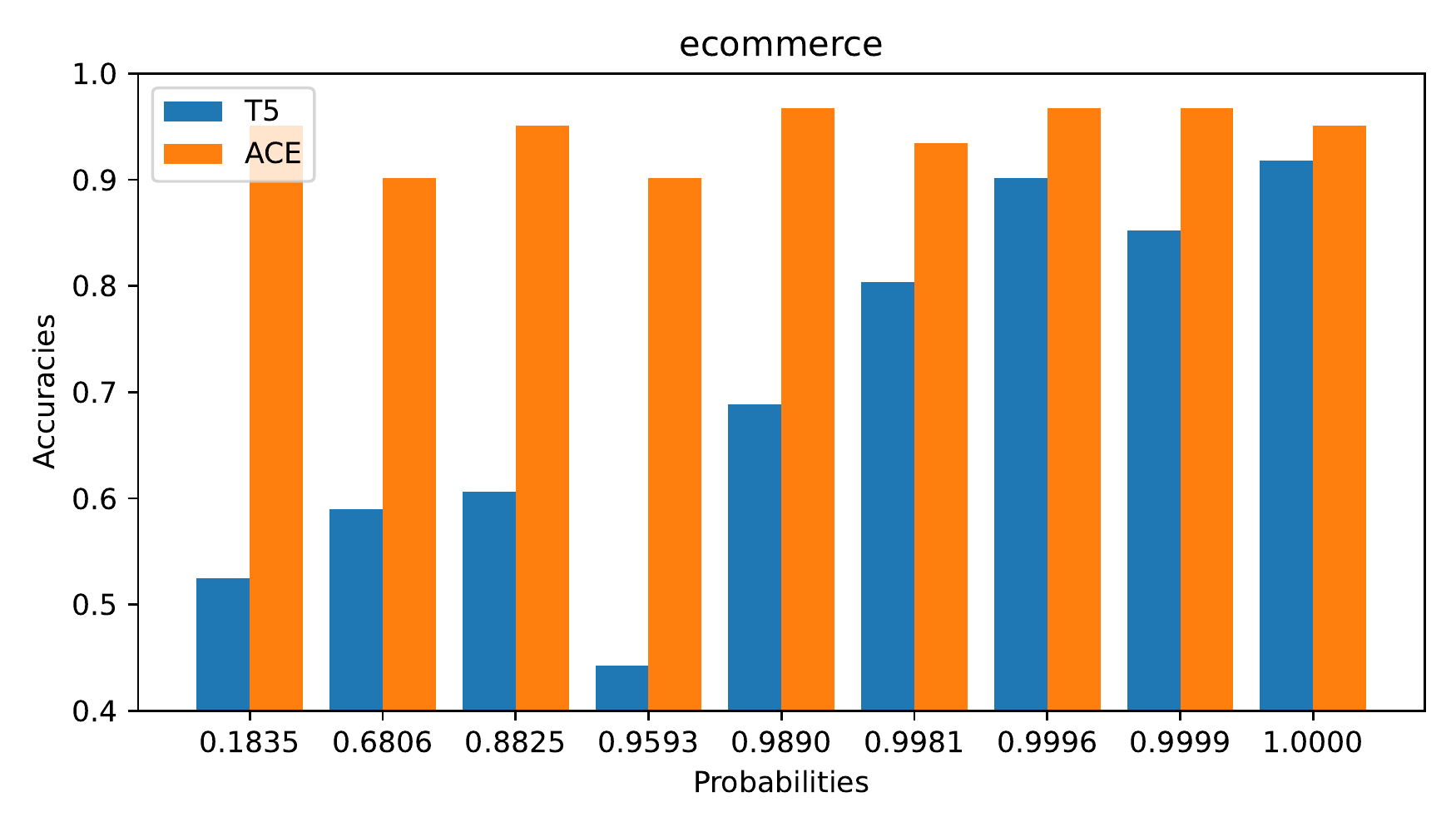} 
 }\\\vspace{-1.2em}
  \subfigure{
    \includegraphics[width=0.4\linewidth]{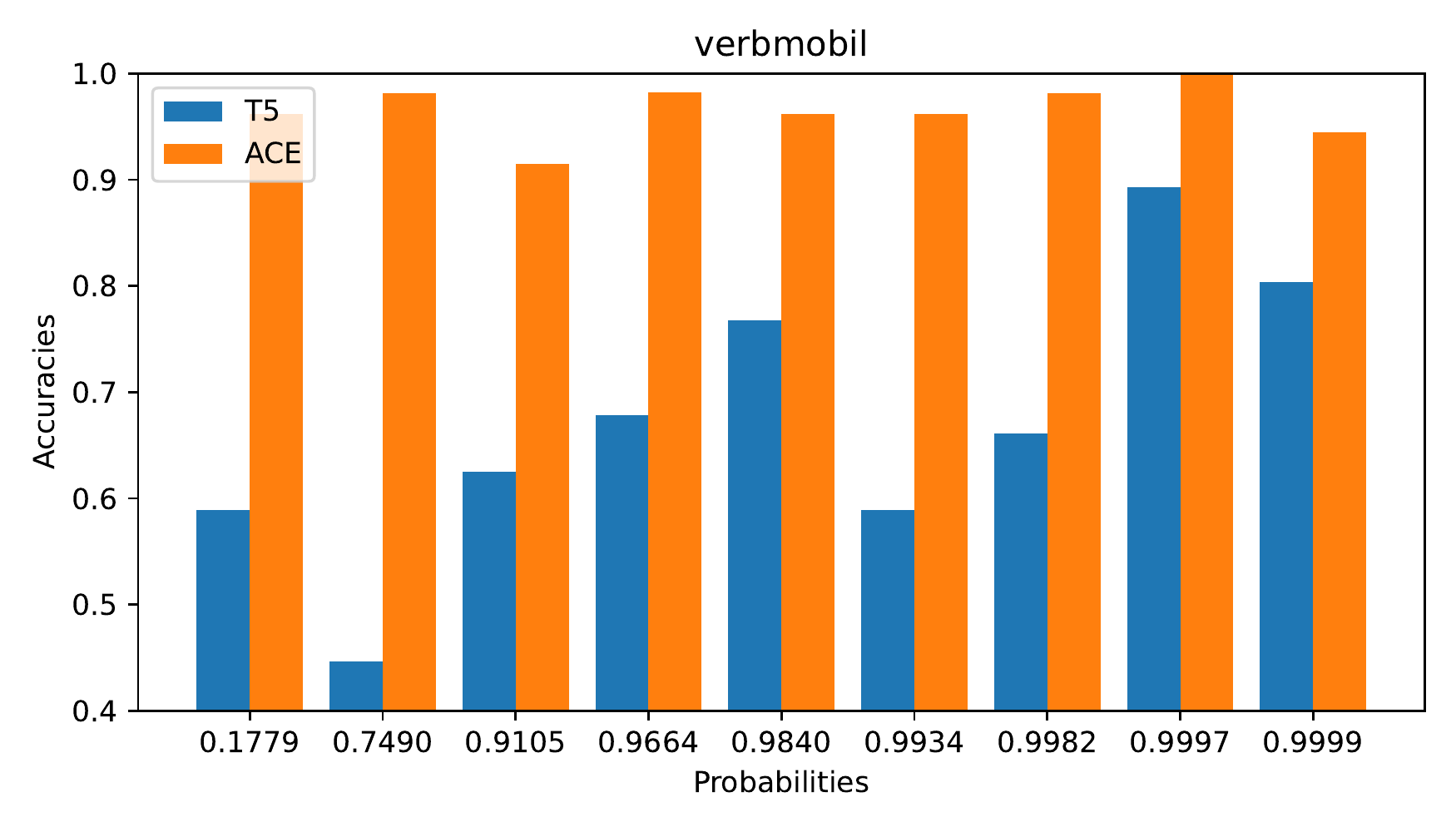} 
 }
  \subfigure{
    \includegraphics[width=0.4\linewidth]{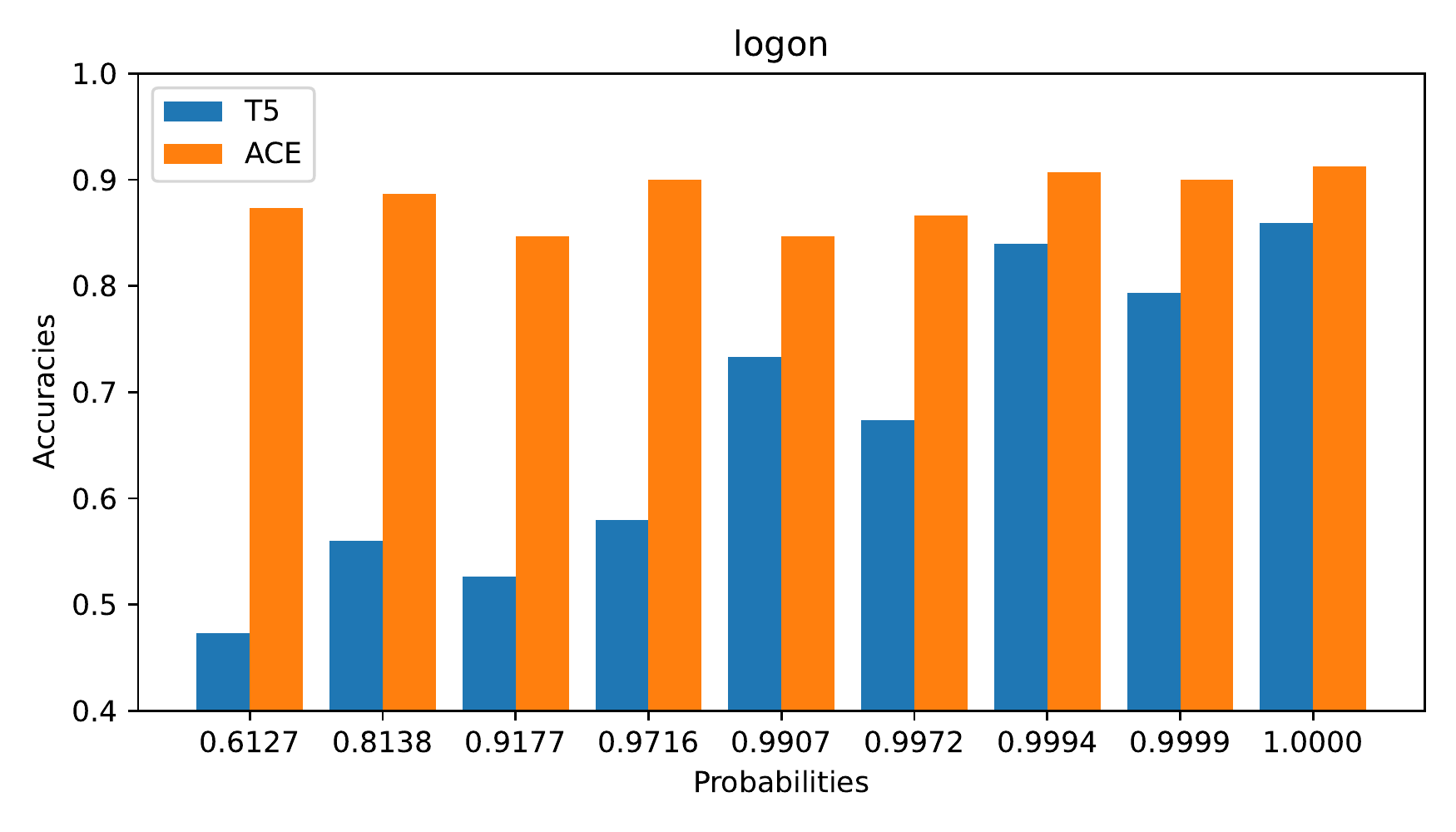} 
 }
 \vspace{-1.5em}
\caption{Diagrams for the T5 model's probabilities verses the T5 model's and ACE parser's accuracies at subgraph level on the other datasets. Each bin contains the same number of examples. Since at most of the subgraphs, the model is pretty certain ($\log P > -1e-5$), we exclude these pretty certain predictions in the figures.}
\label{fig:calibration-other}
\end{figure*}

\section{In-domain Evaluation}
\label{app:ind-test}
\input{sections/tables/ind-perf}
Table~\ref{tab:smatch_tst} shows the in-domain performance, where we compare our parser with the grammar-based ACE parser and other data-driven parsers. The baseline models also include a similar practice with \cite{shaw-etal-2021-compositional} and \citep{hoang2021ensembling}. The former one takes T5 as a backup for grammar-based parser (ACE), and the latter gets ensembled graph via a voting strategy based on the candidates from the T5 parser and ACE parser.

From the table we can see that our methods outperforms the base model (T5-based) and most of the previous work. Specifically, we achieves a \textsc{Smatch} score of $96.77$, which is a $6.11\%$ error reduction compared to the base T5 parser.

%% file: sections/tables/ind-perf.tex
\begin{table}[ht]
  \centering
  \resizebox{\columnwidth}{!}{
  \begin{tabular}{l|ccc}
    \toprule
     \textbf{Model}	&  \textbf{Node} & \textbf{Edge} & \textbf{\textsc{Smatch}} \\
    \midrule
    ACE & 89.30 & 85.05 & 87.14\\
    ACE*  & 93.18 & 88.76 & 90.94 \\
    \citet{buys-blunsom-2017-robust}  & 89.06  & 84.96 & 87.00 \\
    \citet{chen-etal-2018-accurate} & 94.51  & 87.29 & 90.86 \\ 
    \citet{chen-etal-2019-peking} & 95.63 & 91.43 & 93.56 \\
    \citet{chen-etal-2019-peking} & 97.28 & 94.03 & 95.67\\
    \citet{cao-etal-2021-comparing} & 96.42  & 93.73  & 95.05 \\
    ACE-T5 (following \citet{shaw-etal-2021-compositional}) & 93.46 & 89.19 & 91.30\\
    \midrule
    T5-based \citep{lin-etal-2022-towards} & 97.34 & 95.80 & 96.56 \\
    + \citet{hoang2021ensembling} & 88.89 &  87.67 & 88.22\\
    + \citet{lin-etal-2022-towards} & \textbf{97.64} & \textbf{96.41} & \textbf{97.01}\\
    + Ours & 97.50 & 96.07 & 96.77\\
    \bottomrule
  \end{tabular}}
  \caption{F1 score for node and edge predictions and the \textsc{Smatch} scores on the in-domain test set. ACE* refers to evaluation results only for valid parse.}
  \label{tab:smatch_tst}
\end{table}

%% file: sections/appendices/linguistic-phenomena.tex
\section{Fine-grained Linguistic Phenomena}
\label{app:fine-grained-phenomena}
\paragraph{Lexical construction} ERG uses the abstract node {\tt compound} to denote compound words. The edge labeled with {\tt ARG1} refers to the root of the compound word, and thus can help to further distinguish the type of the compound into (1) nominal with normalization, e.g., ``flag burning''; (2) nominal with noun, e.g., ``pilot union''; (3) verbal, e.g., ``state-owned''; (4) named entities, e.g., ``West Germany''.

\paragraph{Argument structure} In ERG, there are different types of core predicates in argument structures, specifically, verbs, nouns and adjectives. We also categorize verb in to basic verb (e.g., {\tt \_look\_v\_1}) and verb particle constructions (e.g., {\tt \_look\_v\_up}). The verb particle construction is handled semantically by having the verb contribute a relation particular to the combination.

\paragraph{Coreference} ERG resolves sentence-level coreference, i.e., if the sentence referring to the same entity, the entity will be an argument for all the nodes that it is an argument of, e.g., in the sentence, ``What we want to do is take a more aggressive stance'', the predicates ``want'' ({\tt \_want\_v\_1}) and ``take'' ({\tt \_take\_v\_1}) share the same agent ``we'' ({\tt pron}).
Coreference can be presented as reentrancies in the ERG graph, we notice that one important type of reentrancies is the passive construction, so we also report evaluation on passive construction in Table \ref{tab:fine-grained}.

%% file: sections/appendices/calibration.tex
\section{Calibration Performance on Other Datasets}
\label{app:calibration}
The correlations between the subgraph's probability and performance on other datasets are shown in Figure \ref{fig:calibration-other}. The conclusions drew from the figure is similar to the one discussed in Section \ref{sec:calibration}.

%% file: emnlp2022.bbl
\begin{thebibliography}{47}
\expandafter\ifx\csname natexlab\endcsname\relax\def\natexlab#1{#1}\fi

\bibitem[{Abend and Rappoport(2013)}]{abend-rappoport-2013-universal}
Omri Abend and Ari Rappoport. 2013.
\newblock \href {https://aclanthology.org/P13-1023} {{U}niversal {C}onceptual
  {C}ognitive {A}nnotation ({UCCA})}.
\newblock In \emph{Proceedings of the 51st Annual Meeting of the Association
  for Computational Linguistics (Volume 1: Long Papers)}, pages 228--238,
  Sofia, Bulgaria. Association for Computational Linguistics.

\bibitem[{Banarescu et~al.(2013)Banarescu, Bonial, Cai, Georgescu, Griffitt,
  Hermjakob, Knight, Koehn, Palmer, and
  Schneider}]{banarescu-etal-2013-abstract}
Laura Banarescu, Claire Bonial, Shu Cai, Madalina Georgescu, Kira Griffitt, Ulf
  Hermjakob, Kevin Knight, Philipp Koehn, Martha Palmer, and Nathan Schneider.
  2013.
\newblock \href {https://aclanthology.org/W13-2322} {{A}bstract {M}eaning
  {R}epresentation for sembanking}.
\newblock In \emph{Proceedings of the 7th Linguistic Annotation Workshop and
  Interoperability with Discourse}, pages 178--186, Sofia, Bulgaria.
  Association for Computational Linguistics.

\bibitem[{Barzdins and Gosko(2016)}]{barzdins-gosko-2016-riga}
Guntis Barzdins and Didzis Gosko. 2016.
\newblock \href {https://doi.org/10.18653/v1/S16-1176} {{RIGA} at
  {S}em{E}val-2016 task 8: Impact of {S}match extensions and character-level
  neural translation on {AMR} parsing accuracy}.
\newblock In \emph{Proceedings of the 10th International Workshop on Semantic
  Evaluation ({S}em{E}val-2016)}, pages 1143--1147, San Diego, California.
  Association for Computational Linguistics.

\bibitem[{Bos et~al.(2004)Bos, Clark, Steedman, Curran, and
  Hockenmaier}]{bos2004wide}
Johan Bos, Stephen Clark, Mark Steedman, James~R Curran, and Julia Hockenmaier.
  2004.
\newblock Wide-coverage semantic representations from a ccg parser.
\newblock In \emph{COLING 2004: Proceedings of the 20th International
  Conference on Computational Linguistics}, pages 1240--1246.

\bibitem[{Buys and Blunsom(2017)}]{buys-blunsom-2017-robust}
Jan Buys and Phil Blunsom. 2017.
\newblock \href {https://doi.org/10.18653/v1/P17-1112} {Robust incremental
  neural semantic graph parsing}.
\newblock In \emph{Proceedings of the 55th Annual Meeting of the Association
  for Computational Linguistics (Volume 1: Long Papers)}, pages 1215--1226,
  Vancouver, Canada. Association for Computational Linguistics.

\bibitem[{Cai and Knight(2013)}]{cai-knight-2013-smatch}
Shu Cai and Kevin Knight. 2013.
\newblock \href {https://aclanthology.org/P13-2131} {{S}match: an evaluation
  metric for semantic feature structures}.
\newblock In \emph{Proceedings of the 51st Annual Meeting of the Association
  for Computational Linguistics (Volume 2: Short Papers)}, pages 748--752,
  Sofia, Bulgaria. Association for Computational Linguistics.

\bibitem[{Callmeier(2000)}]{callmeier2000pet}
Ulrich Callmeier. 2000.
\newblock Pet--a platform for experimentation with efficient hpsg processing
  techniques.
\newblock \emph{Natural Language Engineering}, 6(1):99--107.

\bibitem[{Cao et~al.(2021)Cao, Lin, Sun, and Wan}]{cao-etal-2021-comparing}
Junjie Cao, Zi~Lin, Weiwei Sun, and Xiaojun Wan. 2021.
\newblock \href {https://doi.org/10.1162/coli_a_00395} {Comparing
  knowledge-intensive and data-intensive models for {E}nglish resource semantic
  parsing}.
\newblock \emph{Computational Linguistics}, 47(1):43--68.

\bibitem[{Chen et~al.(2018)Chen, Sun, and Wan}]{chen-etal-2018-accurate}
Yufei Chen, Weiwei Sun, and Xiaojun Wan. 2018.
\newblock \href {https://doi.org/10.18653/v1/P18-1038} {Accurate {SHRG}-based
  semantic parsing}.
\newblock In \emph{Proceedings of the 56th Annual Meeting of the Association
  for Computational Linguistics (Volume 1: Long Papers)}, pages 408--418,
  Melbourne, Australia. Association for Computational Linguistics.

\bibitem[{Chen et~al.(2019)Chen, Ye, and Sun}]{chen-etal-2019-peking}
Yufei Chen, Yajie Ye, and Weiwei Sun. 2019.
\newblock \href {https://doi.org/10.18653/v1/K19-2016} {Peking at {MRP} 2019:
  Factorization- and composition-based parsing for elementary dependency
  structures}.
\newblock In \emph{Proceedings of the Shared Task on Cross-Framework Meaning
  Representation Parsing at the 2019 Conference on Natural Language Learning},
  pages 166--176, Hong Kong. Association for Computational Linguistics.

\bibitem[{Cheng et~al.(2019)Cheng, Reddy, Saraswat, and
  Lapata}]{cheng-etal-2019-learning}
Jianpeng Cheng, Siva Reddy, Vijay Saraswat, and Mirella Lapata. 2019.
\newblock \href {https://doi.org/10.1162/coli_a_00342} {Learning an executable
  neural semantic parser}.
\newblock \emph{Computational Linguistics}, 45(1):59--94.

\bibitem[{Cole et~al.(2021)Cole, Jiang, Pasupat, He, and
  Shaw}]{cole-etal-2021-graph-based}
Jeremy Cole, Nanjiang Jiang, Panupong Pasupat, Luheng He, and Peter Shaw. 2021.
\newblock \href {https://doi.org/10.18653/v1/2021.findings-emnlp.341}
  {Graph-based decoding for task oriented semantic parsing}.
\newblock In \emph{Findings of the Association for Computational Linguistics:
  EMNLP 2021}, pages 4057--4065, Punta Cana, Dominican Republic. Association
  for Computational Linguistics.

\bibitem[{Copestake(2009)}]{copestake-2009-invited}
Ann Copestake. 2009.
\newblock \href {https://aclanthology.org/E09-1001} {\textbf{Invited Talk:}
  slacker semantics: Why superficiality, dependency and avoidance of commitment
  can be the right way to go}.
\newblock In \emph{Proceedings of the 12th Conference of the {E}uropean Chapter
  of the {ACL} ({EACL} 2009)}, pages 1--9, Athens, Greece. Association for
  Computational Linguistics.

\bibitem[{Copestake et~al.(2005)Copestake, Flickinger, Pollard, and
  Sag}]{copestake2005minimal}
Ann Copestake, Dan Flickinger, Carl Pollard, and Ivan~A Sag. 2005.
\newblock Minimal recursion semantics: An introduction.
\newblock \emph{Research on language and computation}, 3(2):281--332.

\bibitem[{Crysmann and Packard(2012)}]{crysmann-packard-2012-towards}
Berthold Crysmann and Woodley Packard. 2012.
\newblock \href {https://aclanthology.org/C12-1043} {Towards efficient {HPSG}
  generation for {G}erman, a non-configurational language}.
\newblock In \emph{Proceedings of {COLING} 2012}, pages 695--710, Mumbai,
  India. The COLING 2012 Organizing Committee.

\bibitem[{Flickinger et~al.(2014)Flickinger, Bender, and
  Oepen}]{flickinger-etal-2014-towards}
Dan Flickinger, Emily~M. Bender, and Stephan Oepen. 2014.
\newblock \href
  {http://www.lrec-conf.org/proceedings/lrec2014/pdf/562_Paper.pdf} {Towards an
  encyclopedia of compositional semantics: Documenting the interface of the
  {E}nglish {R}esource {G}rammar}.
\newblock In \emph{Proceedings of the Ninth International Conference on
  Language Resources and Evaluation ({LREC}'14)}, pages 875--881, Reykjavik,
  Iceland. European Language Resources Association (ELRA).

\bibitem[{Gal and Ghahramani(2016)}]{gal2016dropout}
Yarin Gal and Zoubin Ghahramani. 2016.
\newblock Dropout as a bayesian approximation: Representing model uncertainty
  in deep learning.
\newblock In \emph{international conference on machine learning}, pages
  1050--1059. PMLR.

\bibitem[{Green and {\v{Z}}abokrtsk{\'y}(2012)}]{green-zabokrtsky-2012-hybrid}
Nathan Green and Zden{\v{e}}k {\v{Z}}abokrtsk{\'y}. 2012.
\newblock \href {https://aclanthology.org/W12-0503} {Hybrid combination of
  constituency and dependency trees into an ensemble dependency parser}.
\newblock In \emph{Proceedings of the Workshop on Innovative Hybrid Approaches
  to the Processing of Textual Data}, pages 19--26, Avignon, France.
  Association for Computational Linguistics.

\bibitem[{Hoang et~al.(2021)Hoang, Picco, Hou, Lee, Nguyen, Phan, L{\'o}pez,
  and Fernandez~Astudillo}]{hoang2021ensembling}
Thanh~Lam Hoang, Gabriele Picco, Yufang Hou, Young-Suk Lee, Lam Nguyen, Dzung
  Phan, Vanessa L{\'o}pez, and Ramon Fernandez~Astudillo. 2021.
\newblock Ensembling graph predictions for amr parsing.
\newblock \emph{Advances in Neural Information Processing Systems},
  34:8495--8505.

\bibitem[{H{\"u}llermeier and Waegeman(2021)}]{hullermeier2021aleatoric}
Eyke H{\"u}llermeier and Willem Waegeman. 2021.
\newblock Aleatoric and epistemic uncertainty in machine learning: An
  introduction to concepts and methods.
\newblock \emph{Machine Learning}, 110(3):457--506.

\bibitem[{Ivanova et~al.(2012)Ivanova, Oepen, {\O}vrelid, and
  Flickinger}]{ivanova-etal-2012-contrastive}
Angelina Ivanova, Stephan Oepen, Lilja {\O}vrelid, and Dan Flickinger. 2012.
\newblock \href {https://aclanthology.org/W12-3602} {Who did what to whom? a
  contrastive study of syntacto-semantic dependencies}.
\newblock In \emph{Proceedings of the Sixth Linguistic Annotation Workshop},
  pages 2--11, Jeju, Republic of Korea. Association for Computational
  Linguistics.

\bibitem[{Kasper(1989)}]{kasper1989flexible}
Robert~T Kasper. 1989.
\newblock A flexible interface for linking applications to penman’s sentence
  generator.
\newblock In \emph{Speech and Natural Language: Proceedings of a Workshop Held
  at Philadelphia, Pennsylvania, February 21-23, 1989}.

\bibitem[{Kim(2021)}]{kim2021sequence}
Yoon Kim. 2021.
\newblock Sequence-to-sequence learning with latent neural grammars.
\newblock \emph{Advances in Neural Information Processing Systems},
  34:26302--26317.

\bibitem[{Konstas et~al.(2017)Konstas, Iyer, Yatskar, Choi, and
  Zettlemoyer}]{konstas-etal-2017-neural}
Ioannis Konstas, Srinivasan Iyer, Mark Yatskar, Yejin Choi, and Luke
  Zettlemoyer. 2017.
\newblock \href {https://doi.org/10.18653/v1/P17-1014} {Neural {AMR}:
  Sequence-to-sequence models for parsing and generation}.
\newblock In \emph{Proceedings of the 55th Annual Meeting of the Association
  for Computational Linguistics (Volume 1: Long Papers)}, pages 146--157,
  Vancouver, Canada. Association for Computational Linguistics.

\bibitem[{LeBrun et~al.(2022)LeBrun, Sordoni, and
  O'Donnell}]{lebrun2022evaluating}
Benjamin LeBrun, Alessandro Sordoni, and Timothy~J. O'Donnell. 2022.
\newblock \href {https://openreview.net/forum?id=bTteFbU99ye} {Evaluating
  distributional distortion in neural language modeling}.
\newblock In \emph{International Conference on Learning Representations}.

\bibitem[{Lin et~al.(2022)Lin, Liu, and Shang}]{lin-etal-2022-towards}
Zi~Lin, Jeremiah~Zhe Liu, and Jingbo Shang. 2022.
\newblock \href {https://doi.org/10.18653/v1/2022.findings-acl.328} {Towards
  collaborative neural-symbolic graph semantic parsing via uncertainty}.
\newblock In \emph{Findings of the Association for Computational Linguistics:
  ACL 2022}, pages 4160--4173, Dublin, Ireland. Association for Computational
  Linguistics.

\bibitem[{Liu et~al.(2020)Liu, Lin, Padhy, Tran, Bedrax~Weiss, and
  Lakshminarayanan}]{liu2020simple}
Jeremiah Liu, Zi~Lin, Shreyas Padhy, Dustin Tran, Tania Bedrax~Weiss, and
  Balaji Lakshminarayanan. 2020.
\newblock Simple and principled uncertainty estimation with deterministic deep
  learning via distance awareness.
\newblock \emph{Advances in Neural Information Processing Systems},
  33:7498--7512.

\bibitem[{Malinin and Gales(2020)}]{malinin2020uncertainty}
Andrey Malinin and Mark Gales. 2020.
\newblock Uncertainty estimation in autoregressive structured prediction.
\newblock In \emph{International Conference on Learning Representations}.

\bibitem[{McDonald(2006)}]{mcdonald2006discriminative}
Ryan McDonald. 2006.
\newblock \emph{Discriminative learning and spanning tree algorithms for
  dependency parsing}.
\newblock University of Pennsylvania Philadelphia.

\bibitem[{Murphy(2012)}]{murphy2012machine}
Kevin~P Murphy. 2012.
\newblock \emph{Machine learning: a probabilistic perspective}.
\newblock MIT press.

\bibitem[{Nivre(2008)}]{nivre-2008-algorithms}
Joakim Nivre. 2008.
\newblock \href {https://doi.org/10.1162/coli.07-056-R1-07-027} {Algorithms for
  deterministic incremental dependency parsing}.
\newblock \emph{Computational Linguistics}, 34(4):513--553.

\bibitem[{Oepen et~al.(2020)Oepen, Abend, Abzianidze, Bos, Hajic, Hershcovich,
  Li, O{'}Gorman, Xue, and Zeman}]{oepen-etal-2020-mrp}
Stephan Oepen, Omri Abend, Lasha Abzianidze, Johan Bos, Jan Hajic, Daniel
  Hershcovich, Bin Li, Tim O{'}Gorman, Nianwen Xue, and Daniel Zeman. 2020.
\newblock \href {https://doi.org/10.18653/v1/2020.conll-shared.1} {{MRP} 2020:
  The second shared task on cross-framework and cross-lingual meaning
  representation parsing}.
\newblock In \emph{Proceedings of the CoNLL 2020 Shared Task: Cross-Framework
  Meaning Representation Parsing}, pages 1--22, Online. Association for
  Computational Linguistics.

\bibitem[{Oepen et~al.(2019)Oepen, Abend, Hajic, Hershcovich, Kuhlmann,
  O{'}Gorman, Xue, Chun, Straka, and Uresova}]{oepen-etal-2019-mrp}
Stephan Oepen, Omri Abend, Jan Hajic, Daniel Hershcovich, Marco Kuhlmann, Tim
  O{'}Gorman, Nianwen Xue, Jayeol Chun, Milan Straka, and Zdenka Uresova. 2019.
\newblock \href {https://doi.org/10.18653/v1/K19-2001} {{MRP} 2019:
  Cross-framework meaning representation parsing}.
\newblock In \emph{Proceedings of the Shared Task on Cross-Framework Meaning
  Representation Parsing at the 2019 Conference on Natural Language Learning},
  pages 1--27, Hong Kong. Association for Computational Linguistics.

\bibitem[{Oepen and Flickinger(2019)}]{oepen-flickinger-2019-erg}
Stephan Oepen and Dan Flickinger. 2019.
\newblock \href {https://doi.org/10.18653/v1/K19-2003} {The {ERG} at {MRP}
  2019: Radically compositional semantic dependencies}.
\newblock In \emph{Proceedings of the Shared Task on Cross-Framework Meaning
  Representation Parsing at the 2019 Conference on Natural Language Learning},
  pages 40--44, Hong Kong. Association for Computational Linguistics.

\bibitem[{Oepen et~al.(2015)Oepen, Kuhlmann, Miyao, Zeman, Cinkov{\'a},
  Flickinger, Haji{\v{c}}, and Ure{\v{s}}ov{\'a}}]{oepen-etal-2015-semeval}
Stephan Oepen, Marco Kuhlmann, Yusuke Miyao, Daniel Zeman, Silvie Cinkov{\'a},
  Dan Flickinger, Jan Haji{\v{c}}, and Zde{\v{n}}ka Ure{\v{s}}ov{\'a}. 2015.
\newblock \href {https://doi.org/10.18653/v1/S15-2153} {{S}em{E}val 2015 task
  18: Broad-coverage semantic dependency parsing}.
\newblock In \emph{Proceedings of the 9th International Workshop on Semantic
  Evaluation ({S}em{E}val 2015)}, pages 915--926, Denver, Colorado. Association
  for Computational Linguistics.

\bibitem[{Oepen and L{\o}nning(2006)}]{oepen-lonning-2006-discriminant}
Stephan Oepen and Jan~Tore L{\o}nning. 2006.
\newblock \href {http://www.lrec-conf.org/proceedings/lrec2006/pdf/364_pdf.pdf}
  {Discriminant-based {MRS} banking}.
\newblock In \emph{Proceedings of the Fifth International Conference on
  Language Resources and Evaluation ({LREC}{'}06)}, Genoa, Italy. European
  Language Resources Association (ELRA).

\bibitem[{Oepen et~al.(2002)Oepen, Toutanova, Shieber, Manning, Flickinger, and
  Brants}]{oepen-etal-2002-lingo}
Stephan Oepen, Kristina Toutanova, Stuart Shieber, Christopher Manning, Dan
  Flickinger, and Thorsten Brants. 2002.
\newblock \href {https://aclanthology.org/C02-2025} {The {L}in{GO} redwoods
  treebank: Motivation and preliminary applications}.
\newblock In \emph{{COLING} 2002: The 17th International Conference on
  Computational Linguistics: Project Notes}.

\bibitem[{Ott et~al.(2018)Ott, Auli, Grangier, and Ranzato}]{ott2018analyzing}
Myle Ott, Michael Auli, David Grangier, and Marc’Aurelio Ranzato. 2018.
\newblock Analyzing uncertainty in neural machine translation.
\newblock In \emph{International Conference on Machine Learning}, pages
  3956--3965. PMLR.

\bibitem[{Peng and Gildea(2016)}]{peng-gildea-2016-uofr}
Xiaochang Peng and Daniel Gildea. 2016.
\newblock \href {https://doi.org/10.18653/v1/S16-1183} {{U}of{R} at
  {S}em{E}val-2016 task 8: Learning synchronous hyperedge replacement grammar
  for {AMR} parsing}.
\newblock In \emph{Proceedings of the 10th International Workshop on Semantic
  Evaluation ({S}em{E}val-2016)}, pages 1185--1189, San Diego, California.
  Association for Computational Linguistics.

\bibitem[{Peng et~al.(2015)Peng, Song, and Gildea}]{peng-etal-2015-synchronous}
Xiaochang Peng, Linfeng Song, and Daniel Gildea. 2015.
\newblock \href {https://doi.org/10.18653/v1/K15-1004} {A synchronous hyperedge
  replacement grammar based approach for {AMR} parsing}.
\newblock In \emph{Proceedings of the Nineteenth Conference on Computational
  Natural Language Learning}, pages 32--41, Beijing, China. Association for
  Computational Linguistics.

\bibitem[{Peng et~al.(2017)Peng, Wang, Gildea, and
  Xue}]{peng-etal-2017-addressing}
Xiaochang Peng, Chuan Wang, Daniel Gildea, and Nianwen Xue. 2017.
\newblock \href {https://aclanthology.org/E17-1035} {Addressing the data
  sparsity issue in neural {AMR} parsing}.
\newblock In \emph{Proceedings of the 15th Conference of the {E}uropean Chapter
  of the Association for Computational Linguistics: Volume 1, Long Papers},
  pages 366--375, Valencia, Spain. Association for Computational Linguistics.

\bibitem[{Pollard and Sag(1994)}]{pollard1994head}
Carl Pollard and Ivan~A Sag. 1994.
\newblock \emph{Head-driven phrase structure grammar}.
\newblock University of Chicago Press.

\bibitem[{Raffel et~al.(2020)Raffel, Shazeer, Roberts, Lee, Narang, Matena,
  Zhou, Li, and Liu}]{raffel2020exploring}
Colin Raffel, Noam Shazeer, Adam Roberts, Katherine Lee, Sharan Narang, Michael
  Matena, Yanqi Zhou, Wei Li, and Peter~J Liu. 2020.
\newblock Exploring the limits of transfer learning with a unified text-to-text
  transformer.
\newblock \emph{Journal of Machine Learning Research}, 21:1--67.

\bibitem[{Shaw et~al.(2021)Shaw, Chang, Pasupat, and
  Toutanova}]{shaw-etal-2021-compositional}
Peter Shaw, Ming-Wei Chang, Panupong Pasupat, and Kristina Toutanova. 2021.
\newblock \href {https://doi.org/10.18653/v1/2021.acl-long.75} {Compositional
  generalization and natural language variation: Can a semantic parsing
  approach handle both?}
\newblock In \emph{Proceedings of the 59th Annual Meeting of the Association
  for Computational Linguistics and the 11th International Joint Conference on
  Natural Language Processing (Volume 1: Long Papers)}, pages 922--938, Online.
  Association for Computational Linguistics.

\bibitem[{Steedman(2001)}]{steedman2001syntactic}
Mark Steedman. 2001.
\newblock \emph{The syntactic process}.
\newblock MIT press.

\bibitem[{Toutanova et~al.(2005)Toutanova, Manning, Flickinger, and
  Oepen}]{toutanova2005stochastic}
Kristina Toutanova, Christopher~D Manning, Dan Flickinger, and Stephan Oepen.
  2005.
\newblock Stochastic hpsg parse disambiguation using the redwoods corpus.
\newblock \emph{Research on Language and Computation}, 3(1):83--105.

\bibitem[{Yamada and Matsumoto(2003)}]{yamada-matsumoto-2003-statistical}
Hiroyasu Yamada and Yuji Matsumoto. 2003.
\newblock \href {https://aclanthology.org/W03-3023} {Statistical dependency
  analysis with support vector machines}.
\newblock In \emph{Proceedings of the Eighth International Conference on
  Parsing Technologies}, pages 195--206, Nancy, France.

\end{thebibliography}
